\documentclass[10pt,twocolumn]{article}

\usepackage[letterpaper, top=0.75in, bottom=1in, left=0.625in, right=0.625in]{geometry}
\usepackage{times}
\usepackage{authblk}
\usepackage{abstract}

\usepackage{cite}
\usepackage{amsmath,amssymb,amsfonts}
\usepackage{algorithmic}
\usepackage{graphicx}
\usepackage{multirow}
\usepackage{booktabs}
\usepackage{tabularx}
\usepackage{array}
\usepackage[colorlinks=true, linkcolor=black, citecolor=blue, urlcolor=blue]{hyperref}
\usepackage{float}
\usepackage{makecell}
\usepackage{pifont}
\usepackage{algorithm}

\newcolumntype{Y}{>{\centering\arraybackslash}X}
\newcolumntype{C}[1]{>{\centering\arraybackslash}m{#1}}

\usepackage{bm}

\def\BibTeX{{\rm B\kern-.05em{\sc i\kern-.025em b}\kern-.08em
    T\kern-.1667em\lower.7ex\hbox{E}\kern-.125emX}}

\begin{document}

\title{PEEM: Prompt Engineering Evaluation Metrics for Interpretable Joint Evaluation of Prompts and Responses}
\author[1]{Minki Hong}
\author[1]{Eunsoo Lee}
\author[1]{Sohyun Park}
\author[1]{Jihie Kim\thanks{Corresponding author: Jihie Kim (jihie.kim@dgu.edu). This research was supported by the MSIT (Ministry of Science and ICT), Korea, under the ITRC support program (IITP-2026-RS-2020-II201789), and the AI Convergen
ce Innovation Human Resources Development (IITP-2026-RS-2023-00254592) supervised by the IITP.}}
\affil[1]{Department of Computer Science and Artificial Intelligence, Dongguk University, South Korea\\ \break
\texttt{\{jackyh1, dmstn7432\}@dgu.ac.kr, jihie.kim@dgu.edu}}
\date{}

\maketitle

\maketitle

\begin{abstract}
Prompt design is a primary control interface for large language models (LLMs), yet standard evaluations largely reduce performance to answer correctness, obscuring \emph{why} a prompt succeeds or fails and providing little actionable guidance. We propose \textbf{PEEM (Prompt Engineering Evaluation Metrics)}, a unified framework for \emph{joint} and \emph{interpretable} evaluation of both prompts and responses. PEEM defines a structured rubric with \textbf{9 axes}: \textbf{3} prompt criteria (clarity/structure, linguistic quality, fairness) and \textbf{6} response criteria (accuracy, coherence, relevance, objectivity, clarity, conciseness)---and uses an LLM-based evaluator to output (i) scalar scores on a 1--5 Likert scale and (ii) criterion-specific natural-language rationales grounded in the rubric. Across \textbf{7 benchmarks} and \textbf{5 task models}, PEEM’s accuracy axis strongly aligns with conventional accuracy while preserving model rankings (aggregate Spearman $\rho \approx 0.97$, Pearson $r \approx 0.94$, $p<0.001$). A \textbf{multi-evaluator} study with four models shows consistent relative judgments (pairwise $\rho=0.68$--$0.85$), supporting evaluator-agnostic deployment. Beyond alignment, PEEM captures complementary linguistic failure modes and remains informative under prompt perturbations: prompt-quality trends track downstream accuracy under iterative rewrites, semantic adversarial manipulations induce clear score degradation, and meaning-preserving paraphrases yield high stability (robustness rate $\approx$ 76.7--80.6\%). Finally, using only PEEM scores and rationales as feedback, a \textbf{zero-shot} prompt rewriting loop improves downstream accuracy by up to \textbf{11.7} points, outperforming supervised and RL-based prompt-optimization baselines. Overall, PEEM provides a reproducible, criterion-driven protocol that links prompt formulation to response behavior and enables systematic diagnosis and optimization of LLM interactions.
\end{abstract}

\vspace{0.5em}

\section{Introduction}
\label{sec:1}

Large language models (LLMs) have achieved remarkable performance across a wide range of natural language processing (NLP) tasks, driven in large part by advances in prompt engineering~\cite{b1,b2}. 
Prompt design has emerged as a primary control interface for eliciting reasoning behaviors, with techniques such as chain-of-thought (CoT) prompting~\cite{b3} enabling substantial gains in multi-step inference and complex decision making. As a result, differences in prompt formulation can lead to large variations in model behavior even when the underlying model and task remain fixed.

Despite the central role of prompts, existing evaluation practices remain largely output-centric, focusing almost exclusively on final-answer correctness~\cite{b4}. Most benchmarks rely on a single fixed prompt per task, implicitly assuming that model performance is invariant to prompt phrasing~\cite{b5}. This assumption contradicts extensive empirical evidence that LLMs are highly sensitive to prompt structure, wording, and framing~\cite{b6,b13}, raising concerns about fairness, robustness, and reproducibility. Moreover, prompts influence not only whether an answer is correct, but also how it is presented—affecting coherence, clarity, relevance, and objectivity of the response. Ignoring prompt quality therefore introduces a fundamental blind spot in current evaluation methodologies.

\begin{figure}[t]
\centering
\includegraphics[width=\columnwidth]{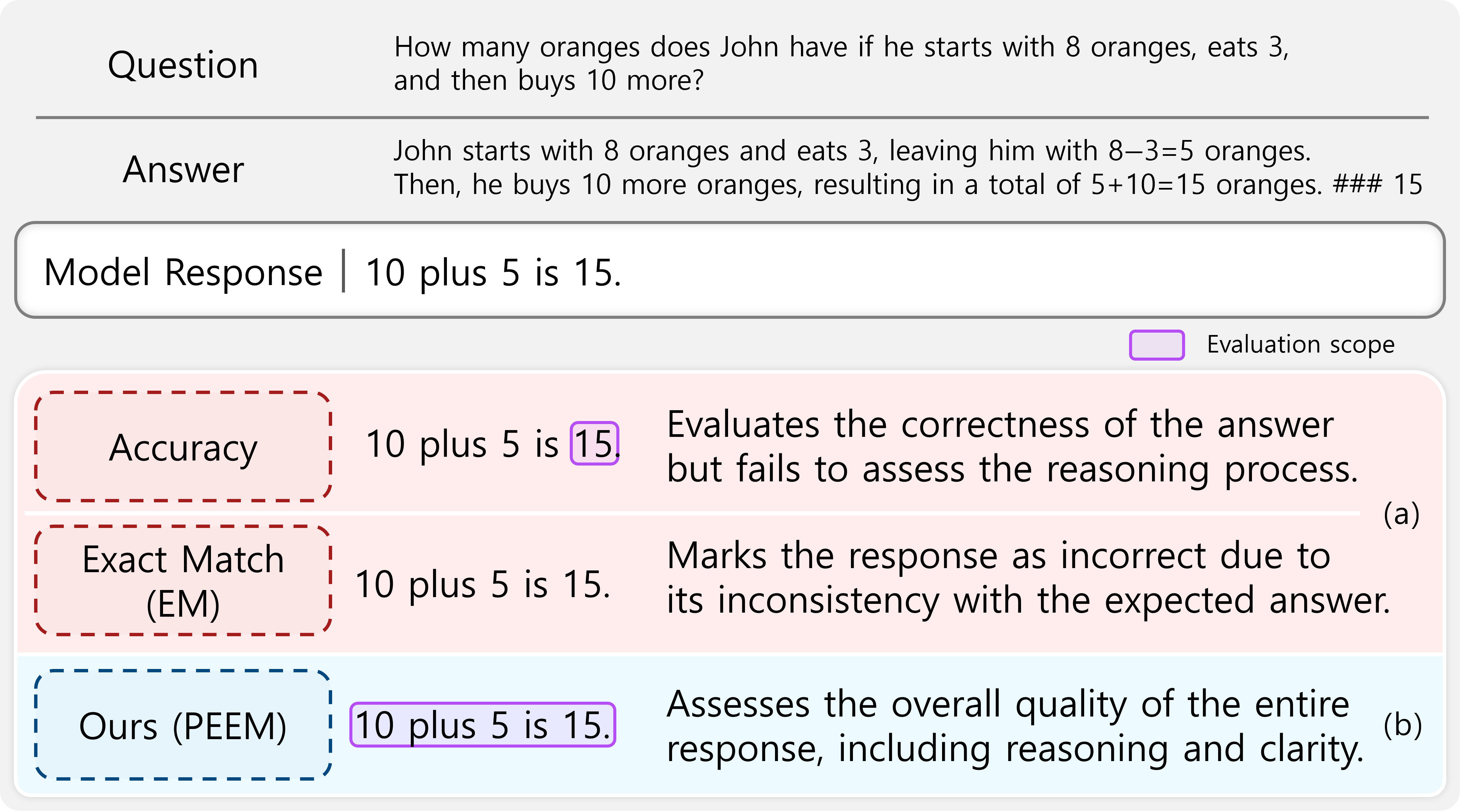}
\caption{Comparison of traditional metrics and our proposed approach, PEEM. \textbf{(a) Accuracy} evaluates only whether the final answer is correct (e.g., ``15"); \textbf{(b) exact match (EM)} requires the response to be identical to the reference answer. Both metrics ignore the reasoning process and linguistic quality. \textbf{(c) PEEM} evaluates the complete response across multiple axes---including logical consistency, contextual relevance, and clarity---while providing interpretable rationales for each score.}
\label{fig:comparison_of_traditional_metrics}
\end{figure}

Traditional metrics such as accuracy and exact match (EM) provide only binary or scalar signals and offer limited visibility into linguistic and semantic dimensions of model behavior (Figure~\ref{fig:comparison_of_traditional_metrics})~\cite{b7,b8}. To address these limitations, recent work has proposed LLM-based evaluators, including GPTScore~\cite{b9} and G-EVAL~\cite{b10}, which assess generated responses along multiple axes. While promising, these methods evaluate \emph{responses in isolation}, typically produce only scalar scores, and provide little interpretable or actionable feedback~\cite{b11}. Crucially, they do not account for how prompt formulation causally shapes response quality, nor do they examine whether such evaluations remain consistent across different evaluator models or under adversarial-style prompt manipulations.

Effective evaluation of LLM systems therefore requires a shift from response-only scoring toward \emph{joint} assessment of prompts and responses, grounded in structured criteria and accompanied by interpretable explanations. Such an evaluation should (i) disentangle prompt quality from response quality, (ii) remain quantitatively aligned with conventional accuracy to preserve comparability and model rankings, (iii) be robust to meaning-preserving paraphrases while remaining sensitive to semantic adversarial-style prompt perturbations, and (iv) demonstrate consistent behavior across multiple evaluator backends without reliance on a single judging model.

In this work, we introduce \textbf{PEEM (Prompt Engineering Evaluation Metrics)}, a unified and interpretable framework for jointly evaluating prompts and responses. PEEM defines a structured rubric consisting of \textbf{nine evaluation axes}: three prompt-level criteria (clarity/structure, linguistic quality, fairness) and six response-level criteria (accuracy, coherence, relevance, objectivity, clarity, conciseness). Given a prompt--response pair, an LLM-based evaluator applies this rubric to generate both (i) scalar scores on a 1--5 Likert scale and (ii) criterion-specific natural-language rationales that justify each score. This dual-mode design enables fine-grained diagnosis of prompt--response interactions beyond surface-level correctness.

We validate PEEM through extensive experiments across seven benchmarks and five task models, using four distinct evaluator models. First, PEEM’s accuracy axis exhibits strong and statistically significant alignment with conventional accuracy while preserving model rankings (aggregate Spearman $\rho \approx 0.97$, Pearson $r \approx 0.94$, $p<0.001$). Second, cross-evaluator analysis shows consistent relative judgments across the tested evaluators, with pairwise rank correlations ranging from $\rho=0.68$ to $0.85$, supporting evaluator-agnostic use in practice. Third, robustness experiments demonstrate that PEEM scores remain stable under meaning-preserving paraphrases, while exhibiting clear and systematic degradation under semantic adversarial-style prompt perturbations, indicating that PEEM captures substantive quality changes rather than superficial lexical variation.

Importantly, PEEM’s rationales are not merely descriptive but \emph{actionable}. Using only PEEM-generated scores and rationales as feedback, we construct a zero-shot prompt rewriting loop that improves downstream accuracy by up to 11.7 points, outperforming supervised and reinforcement learning-based prompt optimization methods. These results show that structured, interpretable evaluation can directly drive prompt optimization without gradient access, additional training data, or human annotation.

In summary, this paper makes the following contributions:
\begin{itemize}
    \item \textbf{We propose PEEM, a unified framework for joint and interpretable evaluation of prompts and responses.} To the best of our knowledge, PEEM is the first framework that integrates prompt-level assessment, multi-axis response evaluation, and criterion-grounded natural-language rationales within a single evaluation protocol.
    \item \textbf{We introduce a robust evaluation paradigm validated under cross-evaluator and adversarial-style settings.} PEEM preserves accuracy-based rankings, remains stable under paraphrasing, and is sensitive to semantic prompt manipulations across multiple tested evaluators.
    \item \textbf{We demonstrate the practical utility of PEEM for prompt optimization.} A zero-shot rewriting loop guided solely by PEEM feedback consistently outperforms supervised and reinforcement learning baselines.
\end{itemize}

\section{Related Work}
\label{sec:2}

\subsection{Evaluation of Prompt Engineering}
\label{sec:2.1}

Prompt engineering has become a central technique for steering the behavior of LLMs, driving substantial performance improvements across a wide range of NLP tasks~\cite{b2,b12}. This progress has sparked growing interest in developing systematic methods for evaluating prompt design. However, most existing work still focuses primarily on downstream accuracy, often overlooking the quality, structure, and role of the prompt itself~\cite{b11}. Recent studies have highlighted that prompt formulation plays a critical role in shaping model outputs~\cite{b6}. For example,~\cite{b13} demonstrates that LLMs are highly sensitive to prompt phrasing, and that performance comparisons across models must account for such variability. Complementary evidence from multimodal settings shows similarly strong prompt sensitivity and proposes model-family-specific prompting principles, underscoring the need for principled prompt evaluation beyond accuracy-only criteria~\cite{b46}.

Despite this growing awareness, response evaluation remains largely dominated by traditional metrics such as accuracy and EM~\cite{b14}. These metrics provide limited coverage and capture only a narrow subset of response quality dimensions~\cite{b5}. As LLMs increasingly generate outputs involving multi-step reasoning, narrative explanations, and subjective judgments, simple correctness-based scoring becomes inadequate~\cite{b11}. To address these limitations, LLM-based evaluation techniques have recently gained traction. G-EVAL~\cite{b10} employs GPT-4 to evaluate generated responses across multiple axes such as coherence, clarity, and relevance, while ChatGPT-as-a-judge~\cite{b15} explores the feasibility of replacing human annotators with LLM judges. While promising, these approaches typically evaluate \emph{responses in isolation}, produce only scalar scores, and lack explicit evaluation criteria or interpretable feedback~\cite{b11}.

Subsequent work has revealed additional limitations of LLM-as-a-judge systems. Prior studies report systematic biases such as position bias and self-preference bias, as well as sensitivity to prompt formatting and surface-level variations in the evaluation prompt itself~\cite{b35,b36,b37,b38,b41}. Beyond English-only settings, cross-lingual automatic evaluation has been explored to assess multilingual LLMs under more diverse linguistic conditions~\cite{b39}. In parallel, prompt-centric research has introduced increasingly sophisticated prompt tuning and selection mechanisms, including uncertainty-aware prompt mixtures (e.g., UMP-Net) and prompt ensembles, to improve robustness and instruction following~\cite{b40}. More recently, evaluation-instructed and evaluation-conditioned frameworks have been proposed to optimize prompts with respect to specific evaluation objectives and query contexts~\cite{b42}. These lines of work collectively underscore the need for evaluation frameworks that are interpretable, robust to prompt variation, and applicable across diverse evaluation settings.

While prior studies address response evaluation, judge reliability, or prompt optimization in isolation, they do not provide a unified framework that jointly evaluates prompts and responses under structured criteria with interpretable rationales. PEEM is designed to fill this specific gap by explicitly modeling prompt--response interactions rather than treating evaluation as a black-box scoring problem applied only to final outputs.

\subsection{Interpretability in Prompt Engineering}
\label{sec:2.2}

Multiple factors in prompt design, including structure, clarity, and stylistic framing, significantly influence the quality of LLM outputs. Recent studies show that even when conveying identical information, variations in prompt phrasing can lead to markedly different responses~\cite{b4}. Nevertheless, most evaluation benchmarks rely on a single fixed prompt per task and offer no mechanism for analyzing prompt sensitivity or interpreting how prompt formulation affects model behavior~\cite{b16,b17}. These limitations have motivated recent efforts to improve the interpretability of prompt evaluation. For instance,~\cite{b18} investigates how different explanation formats affect user trust and understanding of LLM outputs.~Similarly,~\cite{b5} quantitatively analyzes the relationship between prompt formulation and response behavior through controlled multi-prompt experiments. Together, these studies emphasize the importance of examining prompt--response interactions beyond binary correctness.

Complementary robustness-oriented research evaluates how adversarial prompts and red-teaming scenarios can systematically degrade model behavior, highlighting the need for evaluation protocols that remain meaningful under malicious or manipulative prompt conditions~\cite{b43,b44}. However, such work typically focuses on stress-testing model behavior rather than providing interpretable diagnostic signals that can guide prompt refinement.

Our proposed framework, PEEM, directly addresses these challenges by generating criterion-grounded natural language rationales for each evaluation dimension. These rationales enable a feedback loop that supports prompt diagnosis, refinement, and optimization. We empirically validate this capability through a prompt rewriting experiment, demonstrating that interpretable evaluation signals can directly translate into measurable performance improvements.

Table~\ref{tab:comparison_methods} summarizes the key differences between PEEM and existing evaluation approaches, highlighting our contributions in joint prompt--response evaluation and interpretable feedback generation.

\begin{table}[t]
\centering
\caption{Comparison of PEEM with existing LLM evaluation methods.
\textit{Eval.} denotes evaluation; 
\textit{Multi-axis} indicates support for multiple evaluation dimensions; 
\textit{Rationale} denotes explicit justification of evaluation outcomes;
\textit{Actionable} indicates whether feedback can directly guide prompt optimization.}
\label{tab:comparison_methods}
\small
\setlength{\tabcolsep}{2.5pt}
\renewcommand{\arraystretch}{1.20}

\begin{tabularx}{\columnwidth}{@{}
    >{\raggedright\arraybackslash}p{0.26\columnwidth}
    >{\centering\arraybackslash}p{0.12\columnwidth}
    >{\centering\arraybackslash}p{0.14\columnwidth}
    >{\centering\arraybackslash}p{0.12\columnwidth}
    >{\centering\arraybackslash}p{0.12\columnwidth}
    >{\centering\arraybackslash}p{0.12\columnwidth}
@{}}
\toprule
\textbf{Method}
& \textbf{Prompt} \newline \textbf{Eval.}
& \textbf{Response} \newline \textbf{Eval.}
& \textbf{Ratio-} \newline \textbf{nale}
& \textbf{Multi-} \newline \textbf{axis}
& \textbf{Action-} \newline \textbf{able} \\
\midrule
Accuracy/EM
& $\times$ & \checkmark & $\times$ & $\times$ & $\times$ \\
GPTScore~\cite{b9}
& $\times$ & \checkmark & $\times$ & $\times$ & $\times$ \\
G-EVAL~\cite{b10}
& $\times$ & \checkmark & $\times$ & \checkmark & $\times$ \\
ChatGPT-Judge~\cite{b15}
& $\times$ & \checkmark & $\times$ & \checkmark & $\times$ \\
\textbf{PEEM (Ours)}
& \checkmark & \checkmark & \checkmark & \checkmark & \checkmark \\
\bottomrule
\end{tabularx}
\end{table}

\begin{figure*}[t]
    \centering
    \includegraphics[width=\textwidth]{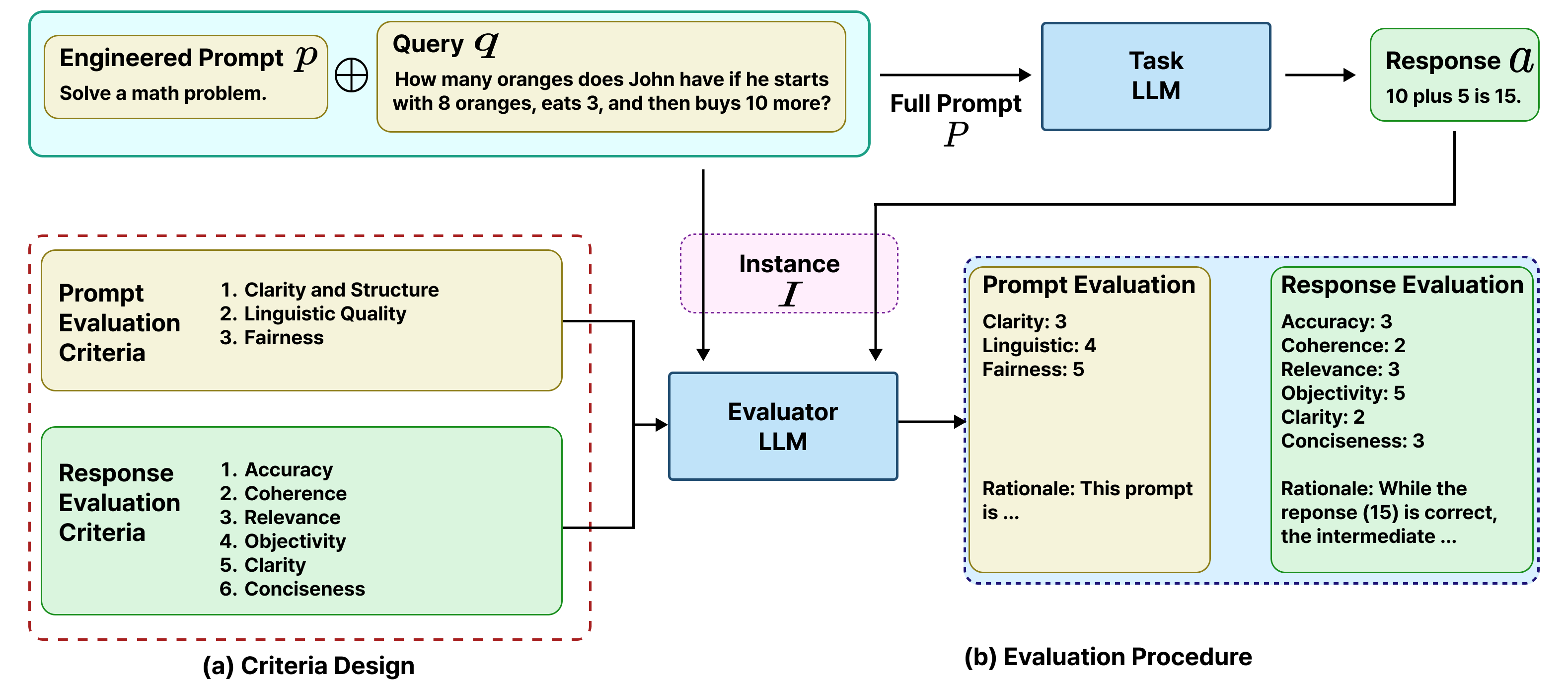}
    \caption{Overview of PEEM. The framework operates in two stages: (a) \textbf{Prompt Evaluation} assesses the input prompt $P = p \oplus q$ across three axes (Clarity, Linguistic Quality, Fairness), and (b) \textbf{Response Evaluation} assesses the model output $a$ across six axes (Accuracy, Coherence, Relevance, Objectivity, Clarity, Conciseness). For each axis, the LLM-based evaluator produces both a numerical score and a natural language rationale explaining the assessment.}
    \label{fig:overall_architecture}
\end{figure*}

\section{Method}
\label{sec:3}
We present the design and implementation of PEEM, a unified framework for evaluating both prompt and response quality in LLM outputs. PEEM performs both quantitative and qualitative assessments, focusing on linguistic properties as well as task alignment. The framework comprises two core components: (1) PEEM Criteria Design and (2) PEEM Evaluation Procedure. 

\paragraph{Preliminaries.}
We clarify key terms used throughout this paper:
\begin{itemize}
    \item \textbf{Zero-shot evaluation}: Assessment without task-specific examples or fine-tuning; the evaluator relies solely on natural language instructions.
    \item \textbf{Prompt~($P = p \oplus q$)}: The concatenation of an engineered instruction $p$ and a task-specific query $q$ (see Eq.~\ref{eq:prompt_concat}).
    \item \textbf{Rationale ($r$)}: A natural language explanation justifying the assigned score for a given evaluation criterion.
    \item \textbf{Multi-axis evaluation}: Simultaneous assessment across multiple independent quality dimensions (e.g., accuracy, coherence, clarity).
    \item \textbf{Task model}: The LLM under evaluation, which receives the prompt $P$ and generates a response $a$. In our experiments, we use five task models: Gemma-2-9B-IT, LLaMA-3.1-8B-IT, Qwen-2.5-7B-IT, GPT-4o-mini, and Gemini-2.5-Flash.
    \item \textbf{Evaluator model}: A separate LLM that applies PEEM's evaluation criteria to assess the quality of the prompt--response pair $(P, a)$. The evaluator produces both scalar scores and natural-language rationales. PEEM is evaluator-agnostic; GPT-4o-mini serves as the default evaluator (see Section~\ref{sec:4.2.1} for cross-evaluator analysis).
\end{itemize}

PEEM uses GPT-4o-mini~\cite{b19} as the primary evaluator model for the following reasons: (1) \textit{Strong instruction-following}: GPT-4o-mini exhibits reliable adherence to structured evaluation templates and produces consistent outputs; (2) \textit{Cost-performance trade-off}: it offers competitive performance at significantly lower API costs (approximately \$0.15 per 1M input tokens), enabling large-scale evaluation across multiple benchmarks; and (3) \textit{Reproducibility}: API versioning helps ensure consistent model behavior across experiments. \textbf{Importantly, this choice is a default implementation decision rather than a methodological requirement.} PEEM is evaluator-agnostic by design: the protocol is defined by criterion-grounded templates and zero-shot instructions, and does not rely on evaluator-specific tuning heuristics. To address concerns about evaluator choice, reproducibility, and sensitivity, we conduct a dedicated \textbf{cross-evaluator analysis} in Section~\ref{sec:4.2.1}, showing that PEEM preserves stable trends and rankings across multiple evaluator models.

The evaluator operates by applying a structured evaluation template that includes scoring prompts and responses across predefined criteria and generating a natural language rationale for each score. Algorithm~\ref{alg:peem} summarizes the overall evaluation procedure, and Figure~\ref{fig:overall_architecture} illustrates the full evaluation pipeline. All evaluations are conducted in a \textit{zero-shot} setting---meaning no task-specific examples, fine-tuning, or in-context demonstrations are provided to the evaluator---ensuring that PEEM generalizes across domains without requiring additional training data. In our experiments, the average evaluation cost per sample was approximately \$0.002, with a processing time of 1.2 seconds per prompt-response pair on average.

\begin{algorithm}[t]
\caption{PEEM Evaluation Procedure}
\label{alg:peem}
\begin{algorithmic}[1]
\REQUIRE Prompt $P = p \oplus q$, Response $a$, Evaluator $\mathcal{M}$
\ENSURE Scores $\{s_i\}$, Rationales $\{r_i\}$
\STATE $C_P \gets \{\text{Clar., Ling., Fair.}\}$ \COMMENT{Prompt criteria}
\STATE $C_R \gets \{\text{Acc., Coh., Rel., Obj., Clar., Conc.}\}$
\STATE $C \gets C_P \cup C_R$ \COMMENT{All 9 criteria}
\FOR{each criterion $c_i \in C$}
    \STATE Build template $T_i$ with rubric for $c_i$
    \STATE $(s_i, r_i) \gets \mathcal{M}(T_i, P, a)$
\ENDFOR
\RETURN $\{(s_i, r_i)\}_{i=1}^{|C|}$
\end{algorithmic}
\end{algorithm}

\subsection{PEEM Criteria Design}
\label{sec:3.1}
\paragraph{Prompt Evaluation Criteria.}
While most prior work in prompt engineering has focused on performance improvement, few studies have proposed systematic criteria for evaluating the quality of the prompt itself~\cite{b20}. Building on foundational principles of prompt design and recent findings on prompt sensitivity~\cite{b13,b17,b46}, we propose a dedicated evaluation schema for prompts, structured around three key axes: Clarity and Structure, Linguistic Quality, and Fairness. Recent work on multimodal prompt sensitivity (Promptception) further motivates principled prompt assessment by showing that small prompt variations can cause substantial performance shifts and by proposing prompting principles tailored to different model families~\cite{b46}. The complete set of evaluation axes for prompt quality is presented in Table~\ref{tab:prompt_criteria}. These criteria aim to ensure that prompts are not only functional but also optimized to align with the diverse capabilities of LLMs. A detailed description of each axis is provided below.

\textbf{Clarity and Structure:} If the intent of the sentence is ambiguous or poorly conveyed, the model may produce outputs that deviate from the task’s objective. This criterion evaluates:
\begin{itemize}
    \item Clear expression of Key Information: The prompt must present all necessary information explicitly and concisely. For example, instead of vague phrasing like ``Analyze this,” the prompt should specify, ``Analyze the sentiment expressed in the following sentence: \{sentence\}.”
    \item Logical Coherence: A well-structured prompt organizes information logically, ensuring the flow of ideas is clear and intuitive. For instance, when multiple pieces of information are provided, they should be ordered to reflect the reasoning steps the model is expected to follow.
\end{itemize}

\begin{table}[t]
\renewcommand{\arraystretch}{1.5} % row spacing
\caption{Prompt evaluation criteria}
\label{tab:prompt_criteria}
\begin{tabularx}{\linewidth}{|p{2cm}|X|}
\hline
\textbf{Criterion} & \textbf{Description} \\ \hline
Clarity and Structure & Assesses the clear articulation of key information, ensuring logical coherence and well-organized presentation. \\ \hline
Linguistic Quality & Evaluates the prompt's linguistic precision, including grammatical accuracy, syntactic coherence, and fluency of expression. \\ \hline
Fairness & Measures the prompt's ability to eliminate biased language and foster inclusivity, ensuring equitable applicability. \\ \hline
\end{tabularx}
\end{table}

\textbf{Linguistic Quality:} The linguistic precision of a prompt directly influences the quality of the generated response. Poor grammar, awkward phrasing, or inconsistencies in tone can reduce the model’s ability to comprehend the task effectively. This criterion evaluates:
\begin{itemize}
    \item Grammatical Accuracy: The prompt must adhere to standard grammatical rules to avoid introducing ambiguity. Errors in sentence structure or verb tense can lead to unintended interpretations.
    \item Fluency and Syntactic Coherence: The prompt should be natural and fluid, using language that aligns with standard conventions. Abrupt or disjointed phrasing may confuse the model.
    \item Domain-Appropriate Language: When the task involves specialized domains (e.g., legal, medical), the prompt should include precise terminology relevant to that field. This helps anchor the model’s responses to the correct context.
    \item Conciseness: While fluency is important, verbose prompts should be avoided to ensure the focus remains on the task.
\end{itemize}

\textbf{Fairness:} LLMs are trained on diverse datasets, which may include inherent biases and stereotypes. Prompts must be designed to mitigate these biases and foster inclusivity. This criterion evaluates:
\begin{itemize}
    \item Minimizing Bias: Prompts should avoid language that perpetuates stereotypes or reinforces societal biases. For example, prompts should not assume gender roles or cultural norms unless explicitly relevant to the task.
    \item Fostering Inclusivity: Prompts must consider a diverse range of users and scenarios. For instance, instead of ``Explain how a businessman handles stress,” a more inclusive phrasing could be, ``Explain how a professional handles stress.”
    \item Neutral and Objective Language: The prompt should maintain neutrality, especially for sensitive or contentious topics. Avoid using emotionally charged or leading language that might skew the model’s response.
\end{itemize}

We define a prompt as the concatenation of two components: an engineered instruction $p$ and a task-defined query $q$, formally expressed as:
\begin{equation}
P = p \oplus q,
\label{eq:prompt_concat}
\end{equation}
where $\oplus$ denotes string concatenation. The goal of prompt engineering is to construct $p$ such that the LLM is guided to generate an appropriate response $a$ to the query $q$~\cite{b2}. Accordingly, our evaluation framework assesses both the intrinsic quality of $p$ and the effectiveness of the composed prompt $P$ in steering model behavior.

\paragraph{Response Evaluation Criteria.}
For response evaluation, we draw inspiration from prior frameworks such as G-EVAL~\cite{b10} and ChatGPT-as-a-Judge~\cite{b15}, and extend them to produce interpretable, criterion-specific scores. Unlike scalar-only methods, PEEM outputs natural language justifications for each axis of evaluation, enabling transparency, error analysis, and prompt refinement.

We define six core evaluation axes: Accuracy, Coherence, Relevance, Objectivity, Clarity, and Conciseness. These axes are selected to balance factual correctness with communicative quality, and each dimension captures a distinct aspect of response utility. Table~\ref{tab:response_criteria} summarizes the criteria, and we elaborate below:

\begin{itemize}
    \item \textbf{Accuracy} measures the factual correctness and logical validity of the response. Responses containing hallucinated claims, reasoning errors, or miscalculations are penalized under this axis.
    
    \item \textbf{Coherence} assesses whether the response is logically structured and maintains a consistent flow across sentences or sections. Disjointed transitions or fragmented ideas degrade the overall coherence.

    \item \textbf{Relevance} evaluates how well the response addresses the specific question or task. Even if the content is accurate, digressions or off-topic elaborations reduce its alignment with the original prompt.

    \item \textbf{Objectivity} determines whether the response is presented in a neutral, unbiased tone. This includes avoiding speculative or emotionally charged language, particularly in ethically or socially sensitive contexts.

    \item \textbf{Clarity} examines whether the response is easy to understand and free from ambiguity. It considers lexical simplicity, syntactic transparency, and the effective communication of intended meaning.

    \item \textbf{Conciseness} judges the ability to convey essential information without redundancy. Responses should be concise while maintaining informativeness and avoiding unnecessary detail.
\end{itemize}

By combining these axes into a unified framework, PEEM supports fine-grained, interpretable evaluation that captures both the semantic fidelity and communicative quality of LLM-generated responses.

\begin{table}[t]
\caption{Response evaluation criteria}
\label{tab:response_criteria}
\renewcommand{\arraystretch}{1.5} % row spacing
\begin{tabularx}{\linewidth}{|p{2cm}|X|}
\hline
\textbf{Criterion} & \textbf{Description} \\ \hline
Accuracy & Measures the correctness of the response, ensuring that factual information and logical conclusions are accurate. \\ \hline
Coherence & Assesses whether the response is logically structured and maintains consistent flow across sentences or sections. \\ \hline
Relevance & Evaluates how well the response addresses the specific question or task, avoiding irrelevant or off-topic content. \\ \hline
Objectivity & Determines whether the response is unbiased, avoids subjective language, and provides balanced perspectives. \\ \hline
Clarity & Examines whether the response is easy to understand, free from ambiguity, and clearly communicates the intended message. \\ \hline
Conciseness & Judges the response’s ability to convey the required information succinctly without unnecessary elaboration. \\ \hline
\end{tabularx}
\end{table}

\begin{table}[t]
\centering
\caption{A sample from response evaluation. PEEM conducts assessments based on the defined criteria. The criteria’s scores and accompanying language rationales offer a broad analysis of the sentence’s content quality and its adherence to the user’s intended requirements. PEEM provides a detailed analysis of the assessment, including these criteria.}
\label{tab:response_example}
\renewcommand{\arraystretch}{1.5} % row spacing
\begin{tabularx}{\columnwidth}{@{}lX@{}}
\toprule
\textbf{Query:} & How many oranges does John have if he starts with 8 oranges, eats 3, and then buys 10 more? \\
\midrule
\textbf{Response:} & 10 plus 5 is 15. \\
\midrule
\multicolumn{2}{@{}l}{\textbf{Response Evaluation}} \\
\end{tabularx}

\renewcommand{\arraystretch}{1.5} % row spacing
\begin{tabularx}{\columnwidth}{@{}lX@{}}
\textbf{Criterion} & \textbf{Score \(s\) \& Language Rationale \(r\)} \\
\midrule
Accuracy    & 3: The final result (15) is correct, but the intermediate calculation (``10 plus 5 is 15") is disconnected from the original problem. \\
Coherence   & 2: The reasoning is disjointed and fails to logically explain how the result was derived. \\
Relevance   & 3: The response partially addresses the question, providing the correct final number but neglecting the problem's context. \\
Objectivity & 5: The response is neutral and fact-based, with no subjective or biased language. \\
Clarity     & 2: Key reasoning steps are omitted, making the response difficult to follow. \\
Conciseness & 3: The response is concise but lacks necessary explanatory details. \\
\bottomrule
\end{tabularx}
\end{table}

%% =========== linked to Sec. 4.2 ===========
\begin{table*}[t]
\setlength{\tabcolsep}{6pt}
\caption{
\textbf{Comparison between traditional accuracy and PEEM-based evaluation across 7 benchmarks and 5 task models.}
\textit{acc.} reports the conventional task accuracy computed against gold labels (in \%).
\textit{PEEM's acc} reports the mean Likert score in \{1,\dots,5\} assigned to the \textbf{Accuracy} axis by the PEEM evaluator, averaged over all instances.
\textit{Response overall} denotes the average of the six response criteria (Accuracy, Coherence, Relevance, Objectivity, Clarity, Conciseness),
and \textit{Prompt overall} denotes the average of the three prompt criteria (Clarity/Structure, Linguistic Quality, Fairness) (cf. Eq.~\ref{eq:prompt_overall}--\ref{eq:response_overall}).
For brevity, benchmark names are abbreviated as ARC-C = ARC-Challenge, ARC-E = ARC-Easy, and BBH = Big-Bench Hard.
Cross-evaluator analysis is provided in Section~\ref{sec:4.2.1}.
}
\label{tab:accuracy_comparison}
\centering
\begin{tabularx}{\textwidth}{@{}l|*{7}{X}@{}}
\toprule
\textbf{Task Model (Metric)} & \textbf{AG News} & \textbf{ARC-C} & \textbf{ARC-E} & \textbf{BBH} & \textbf{GSM8K} & \textbf{MMLU} & \textbf{SST-2} \\
\midrule
Gemma-2-9B-IT (acc.) & 59.0\% & 67.9\% & 72.1\% & 34.9\% & 57.1\% & 67.3\% & 73.9\% \\
Gemma-2-9B-IT (PEEM’s acc) & 4.063 & 4.168 & 4.201 & 3.780 & 4.354 & 4.180 & 4.599 \\
Gemma-2-9B-IT (Response overall) & 4.223 & 4.246 & 4.225 & 4.446 & 4.582 & 4.431 & 4.680 \\
Gemma-2-9B-IT (Prompt overall) & 4.061 & 4.105 & 4.183 & 4.000 & 4.125 & 4.153 & 4.307 \\
\midrule
LLaMA-3.1-8B-IT (acc.) & 55.6\% & 83.4\% & 92.4\% & 58.6\% & 83.6\% & 71.1\% & 90.1\% \\
LLaMA-3.1-8B-IT (PEEM’s acc) & 3.845 & 4.461 & 4.700 & 3.950 & 4.480 & 4.113 & 4.662 \\
LLaMA-3.1-8B-IT (Response overall) & 4.656 & 4.577 & 4.690 & 4.386 & 4.646 & 4.463 & 4.704 \\
LLaMA-3.1-8B-IT (Prompt overall) & 3.958 & 4.350 & 4.484 & 4.106 & 4.405 & 4.252 & 4.451 \\
\midrule
Qwen-2.5-7B-IT (acc.) & 56.3\% & 89.8\% & 95.5\% & 67.6\% & 86.0\% & 68.7\% & 92.2\% \\
Qwen-2.5-7B-IT (PEEM’s acc) & 4.048 & 4.702 & 4.808 & 4.187 & 4.654 & 4.188 & 4.787 \\
Qwen-2.5-7B-IT (Response overall) & 4.750 & 4.692 & 4.745 & 4.537 & 4.723 & 4.577 & 4.714 \\
Qwen-2.5-7B-IT (Prompt overall) & 4.209 & 4.555 & 4.601 & 4.256 & 4.506 & 4.302 & 4.554 \\
\midrule
GPT-4o-mini (acc.) & 85.5\% & 93.7\% & 98.4\% & 78.3\% & 95.3\% & 82.1\% & 93.0\% \\
GPT-4o-mini (PEEM’s acc) & 4.550 & 4.815 & 4.906 & 4.392 & 4.882 & 4.388 & 4.725 \\
GPT-4o-mini (Response overall) & 4.764 & 4.810 & 4.883 & 4.668 & 4.847 & 4.709 & 4.844 \\
GPT-4o-mini (Prompt overall) & 4.457 & 4.702 & 4.809 & 4.354 & 4.701 & 4.500 & 4.753 \\
\midrule
Gemini-2.5-Flash (acc.) & 87.2\% & 91.4\% & 96.8\% & 76.5\% & 94.1\% & 80.3\% & 91.5\% \\
Gemini-2.5-Flash (PEEM's acc) & 4.580 & 4.721 & 4.885 & 4.290 & 4.810 & 4.412 & 4.705 \\
Gemini-2.5-Flash (Response overall) & 4.695 & 4.782 & 4.901 & 4.510 & 4.855 & 4.605 & 4.812 \\
Gemini-2.5-Flash (Prompt overall) & 4.410 & 4.595 & 4.762 & 4.185 & 4.712 & 4.350 & 4.620 \\
\bottomrule
\end{tabularx}
\end{table*}
%% =========== linked to Sec. 4.2 ===========

\subsection{PEEM Evaluation Procedure}
\label{sec:3.2}
Given each prompt-response instance, we define:
\begin{equation}
 I = (P, a),
 \label{eq:instance}
\end{equation}
where $P$ is the full prompt defined in~(\ref{eq:prompt_concat}), and $a$ is the model-generated response.

For every instance $I$ defined in~(\ref{eq:instance}), PEEM generates both a numerical score (on a 1--5 Likert scale) and a natural language rationale based on predefined evaluation criteria. The evaluator model receives a prompt that encodes the full set of evaluation axes and performs an integrated assessment of the instance. Specifically, the composed prompt $P = p \oplus q$ is evaluated along three axes defined in Table~\ref{tab:prompt_criteria}. In comparison, the response $a$ is assessed based on six axes defined in Table~\ref{tab:response_criteria}, as illustrated in Figure~\ref{fig:overall_architecture} (b). For each evaluation axis, the model outputs both a scalar score and a justification that explains the rationale behind the assigned value.

To ensure reproducibility, we explicitly define the aggregated scores used throughout our experiments. Let $C_P$ and $C_R$ denote the prompt and response criteria sets (Algorithm~\ref{alg:peem}). Given criterion-level scores $\{s_c\}$, we compute the prompt and response overall scores as:
\begin{equation}
S_{\text{prompt}}(P) = \frac{1}{|C_P|} \sum_{c \in C_P} s_c,
\label{eq:prompt_overall}
\end{equation}
\begin{equation}
S_{\text{response}}(a) = \frac{1}{|C_R|} \sum_{c \in C_R} s_c.
\label{eq:response_overall}
\end{equation}
In Table~\ref{tab:accuracy_comparison}, ``Prompt overall'' and ``Response overall'' correspond to $S_{\text{prompt}}$ and $S_{\text{response}}$. ``PEEM's acc'' denotes the scalar score assigned to the Accuracy axis within $C_R$.

This approach differs from conventional evaluation methods and offers the following advantages:
\begin{enumerate}
    \item \textbf{Unified quantitative and qualitative assessment:} Scores and rationales are generated in parallel within a single framework, enabling both numerical comparison and interpretable analysis.
    \item \textbf{Integrated prompt--response evaluation:} The evaluator assesses not only the output $a$ but also how the input prompt $P$ shaped the generated response, enabling explicit analysis of prompt-induced effects rather than response-only scoring.
    \item \textbf{Feedback reusability:} Generated rationales can be leveraged for downstream prompt optimization and debugging workflows without additional supervision.
    \item \textbf{Evaluator-agnostic design:} Because PEEM relies on criterion-grounded templates rather than evaluator-specific tuning, the same protocol can be applied across different LLM judges, supporting reproducible and comparable evaluation. This property is empirically validated through the cross-evaluator study in Section~\ref{sec:4.2.1}.
\end{enumerate}
All evaluations are conducted in a zero-shot setting, with no task-specific fine-tuning. The evaluator applies each criterion using only the provided instructions and (\ref{eq:instance}). Examples of PEEM evaluation outputs are provided in Table~\ref{tab:response_example}, and the corresponding evaluation templates are included in Appendix~\ref{appendix:a}.

\section{Experiments}
\label{sec:4}

\subsection{Experimental Setup}
\label{sec:4.1}
To comprehensively and reliably evaluate PEEM, we conduct experiments across a diverse range of benchmarks and model configurations. We use seven datasets spanning classification, reasoning, commonsense inference, and sentiment analysis tasks: AG News~\cite{b21}, ARC-Challenge (ARC-C), ARC-Easy (ARC-E)~\cite{b22}, BigBench-Hard (BBH)~\cite{b23}, GSM8K~\cite{b24}, MMLU~\cite{b25}, and SST-2~\cite{b26}.
For model diversity, we evaluate responses generated by five language models: Gemma-2-9B-IT (instruction-tuned; IT)~\cite{b27}, LLaMA-3.1-8B-IT~\cite{b28}, Qwen-2.5-7B-Instruct (Qwen-2.5-7B-IT)~\cite{b29}, GPT-4o-mini~\cite{b19}, and Gemini-2.5-Flash~\cite{b45}. To assess evaluator robustness, we employ five distinct evaluator models: GPT-4o-mini (default), Gemini-2.5-Flash, Gemma-2-9B-IT, LLaMA-3.1-8B-IT, and Qwen-2.5-7B-IT. This multi-evaluator configuration enables comprehensive analysis of cross-evaluator agreement (Section~\ref{sec:4.2.1}).
We adopt accuracy as a baseline metric for comparison. All experiments are conducted under a zero-shot setting to ensure fair and consistent evaluation. In Section~\ref{sec:4.2}, we use the benchmark-provided engineered prompts when available; otherwise, we default to using the raw query alone.

\subsection{Alignment with Existing Evaluation Metrics}
\label{sec:4.2}

To validate \textsc{PEEM} as a reliable alternative to traditional metrics, we analyze its alignment with accuracy and examine its stability across different evaluator models.

\subsubsection{Correlation with Accuracy}
We analyzed the relationship between \textsc{PEEM}'s accuracy dimension and conventional accuracy across seven benchmarks and five task models (Table~\ref{tab:accuracy_comparison}). Overall, \textsc{PEEM} exhibits strong rank-level alignment with traditional accuracy across both open-weight and proprietary language models, indicating that PEEM-derived scores preserve relative performance ordering while providing richer evaluative signals.

\textsc{LLaMA-3.1-8B-IT} and \textsc{Qwen-2.5-7B-IT} show near-perfect agreement with conventional accuracy ($\rho = 1.00$, $r \approx 0.99$), reflecting highly stable rank ordering across all benchmarks. \textsc{GPT-4o-mini} similarly demonstrates strong correlation ($\rho \approx 0.96$, $r \approx 0.98$). \textsc{Gemini-2.5-Flash} also exhibits strong alignment with conventional accuracy ($\rho \approx 0.96$, $r \approx 1.00$), indicating that improvements in task correctness and response quality are closely coupled for high-performing proprietary models.

In contrast, \textsc{Gemma-2-9B-IT} shows comparatively lower but still positive correlation ($\rho \approx 0.61$, $r \approx 0.77$), reflecting greater variability in reasoning completeness and response consistency. This suggests that correct final answers are not always accompanied by equally robust explanations for mid-scale instruction-tuned models.

The overall correlation is computed across all model--benchmark pairs rather than averaging per-model correlations. While \textsc{Gemma-2-9B-IT} shows moderate within-model variability, the global rank ordering across models is highly consistent, resulting in strong aggregate correlation. Aggregating all models and benchmarks ($n = 35$), we observe an overall Spearman correlation of $\rho \approx 0.97$ and Pearson correlation of $r \approx 0.94$ ($p < 0.001$), confirming strong and statistically significant alignment between PEEM-based accuracy and conventional accuracy despite the limited benchmark coverage.

\subsubsection{Cross-Evaluator Agreement}
\label{sec:4.2.1}
To assess \textsc{PEEM}'s robustness across different evaluator backends, we conducted a multi-evaluator study using four distinct models: Gemma-2-9B-IT, LLaMA-3.1-8B-IT, Qwen-2.5-7B, and Gemini-2.5-Flash. Each evaluator assessed responses from five task models across all seven benchmarks.

Table~\ref{tab:cross_evaluator} summarizes the cross-evaluator agreement measured by Spearman correlation. Open-source evaluators show strong mutual agreement: Gemma-LLaMA ($\rho = 0.82$), LLaMA-Qwen ($\rho = 0.85$), and Gemma-Qwen ($\rho = 0.78$). Gemini maintains moderate-to-strong correlation with all evaluators ($\rho = 0.68$--$0.74$), with variations attributable to its distinct evaluation granularity. Complete results are provided in Appendix~\ref{appendix:c}.

\begin{table}[t]
\centering
\caption{Cross-evaluator Spearman correlation ($\rho$) for PEEM response accuracy scores. Each cell reports the pairwise $\rho$ computed over all task-model $\times$ dataset combinations (5 task models $\times$ 7 benchmarks = 35 score vectors per evaluator). The four evaluator models---Gemini-2.5-Flash, Gemma-2-9B-IT, LLaMA-3.1-8B-IT, and Qwen-2.5-7B-IT---each independently scored the same prompt--response instances using the PEEM evaluation protocol. GPT-4o-mini is used as the default evaluator in all other experiments and is therefore excluded from this pairwise comparison.}
\label{tab:cross_evaluator}
\small
\begin{tabular}{lcccc}
\toprule
& \textbf{Gemini} & \textbf{Gemma} & \textbf{LLaMA} & \textbf{Qwen} \\
\midrule
Gemini & 1.00 & 0.74 & 0.68 & 0.71 \\
Gemma & -- & 1.00 & 0.82 & 0.78 \\
LLaMA & -- & -- & 1.00 & 0.85 \\
Qwen & -- & -- & -- & 1.00 \\
\bottomrule
\end{tabular}
\end{table}

While individual evaluators exhibit characteristic scoring patterns (e.g., Gemma averaging 4.51 vs. Qwen averaging 3.86), the relative ranking of task models remains consistent across all evaluators. This cross-evaluator stability validates \textsc{PEEM}'s generalizability and allows practitioners to select evaluators based on computational constraints without compromising evaluation validity.

Through the experiments in Sections~\ref{sec:4.3} and~\ref{sec:4.4}, we further demonstrate PEEM's utility for prompt optimization. Additional examples are provided in Appendix~\ref{appendix:b}.

\subsubsection{Human Evaluation Alignment}
\label{sec:4.2.2}
To validate PEEM beyond automated correlation and cross-evaluator analyses, we conducted a human evaluation study where three graduate students in artificial intelligence---who were not involved in this research---independently assessed 210 randomly sampled instances (30 per dataset) across the same six response criteria used by PEEM: Accuracy, Coherence, Relevance, Conciseness, Objectivity, and Clarity. Each annotator assigned integer scores on the same 1--5 Likert scale. All annotations were performed on Gemma-2-9B-IT outputs using the PEEM consensus score (averaged across four evaluator models) as the automated baseline. Details of the annotation procedure are provided in Appendix~\ref{appendix:d}.

As shown in Table~\ref{tab:human_eval_criterion}, the overall correlation between PEEM and human judgments is strong, with Pearson $r = 0.84$ and Spearman $\rho = 0.72$ ($p < 0.001$). Per-criterion analysis reveals that Accuracy ($\rho = 0.71$, $r = 0.86$) and Conciseness ($\rho = 0.76$, $r = 0.77$) exhibit the strongest alignment, as these criteria involve relatively unambiguous judgments. Objectivity shows a lower Spearman correlation ($\rho = 0.35$) due to ceiling effects---the majority of samples received near-perfect scores from both PEEM and humans---though the Pearson correlation remains moderate ($r = 0.69$).

Inter-annotator agreement, measured by Krippendorff's $\alpha = 0.59$ and an average pairwise $\pm$1 agreement rate of 99.5\%, indicates consistent human judgments. Notably, human annotators exhibited slightly stricter scoring tendencies (mean $= 4.58$) compared to PEEM (mean $= 4.70$), consistent with prior observations that LLM-based evaluators tend toward lenient scoring~\cite{b11}. Per-dataset alignment and detailed inter-annotator statistics are provided in Appendix~\ref{appendix:d}.

\begin{table}[t]
\centering
\caption{Per-criterion alignment between PEEM and human evaluation. Three annotators independently scored 210 samples across six criteria. All correlations are statistically significant at $p < 0.001$.}
\label{tab:human_eval_criterion}
\small
\begin{tabular}{lcccc}
\toprule
\textbf{Criterion} & \textbf{PEEM} & \textbf{Human} & \textbf{Spearman $\rho$} & \textbf{Pearson $r$} \\
\midrule
Accuracy    & 4.68 & 4.57 & 0.71 & 0.86 \\
Coherence   & 4.77 & 4.62 & 0.55 & 0.72 \\
Relevance   & 4.83 & 4.69 & 0.57 & 0.78 \\
Conciseness & 4.12 & 4.06 & 0.76 & 0.77 \\
Objectivity & 4.94 & 4.81 & 0.35 & 0.69 \\
Clarity     & 4.83 & 4.76 & 0.58 & 0.79 \\
\midrule
\textbf{Overall}    & 4.70 & 4.58 & \textbf{0.72} & \textbf{0.84} \\
\bottomrule
\end{tabular}
\end{table}

%% 4.3에 연관된 표 ====================================================
\begin{table}[t]
\caption{PEEM's rewriting experiment results. The baseline results (top, middle section) are cited from PRewrite~\cite{b30}.}
\label{tab:Comparison with baselines}
\centering
\resizebox{\columnwidth}{!}{
\begin{tabular}{lccc}
\toprule
\textbf{Method} & \textbf{AG News} & \textbf{SST-2} & \textbf{GSM8K} \\
\midrule
AutoPrompt          & 65.7 & 75.0 & -- \\
RLPrompt            & 77.2 & 90.1 & -- \\
TEMPERA             & 81.3 & 92.0 & -- \\
\midrule
Prewrite (Initial Prompt) & 76.9 & 96.3 & 29.9 \\
PRewrite-I                & 84.5 & 96.5 & 52.0 \\
PRewrite-S                & 85.2 & 96.6 & 53.6 \\
\midrule
\textbf{Ours (Gemma-2-9B-IT)} & & & \\
Initial Prompt ($P_i$)     & 59.0 & 73.9 & 57.1 \\
Rewrite ($P_s$)            & 70.4 & 87.7 & 58.8 \\
Rewrite ($P_c$)            & 83.9 & 92.2 & 65.3 \\
\bottomrule
\end{tabular}
}
\end{table}
%% 4.3에 연관된 표 ====================================================

%% 4.3에 연관된 표 (2) ====================================================
\begin{table}[t]
\centering
\caption{Example of prompt rewriting and re-evaluation. This example utilizes the Score with Context \(P_\text{c}\) method to rewrite the prompt based on the rationale \(r\) described in Table~\ref{tab:response_example}. Subsequently, the rewritten prompt \(P_\text{c}\) is used to generate the response \(R_\text{c}\). The re-evaluation results demonstrate that the prompt and response are improved following our rewriting process.}
\label{tab:example_rewriting}
\small
\renewcommand{\arraystretch}{1.5} % 행 간격 조정
\begin{tabularx}{\columnwidth}{@{}lX@{}}
\toprule
\multicolumn{2}{l}{\textbf{Initial Prompt}} \\
$P_i$: & How many oranges does John have if he starts with 8 oranges, eats 3, and then buys 10 more? \\
\midrule
$R_i$: & 10 plus 5 is 15. \\
\bottomrule
\end{tabularx}
\renewcommand{\arraystretch}{1.5} % 행 간격 조정
\begin{tabularx}{\columnwidth}{|X|X|X|X|X|X|}
\multicolumn{2}{@{}l@{}}{\textbf{Response Evaluation}} \\
\hline
\textbf{Acc} & \textbf{Coh} & \textbf{Rel} & \textbf{Obj} & \textbf{Cla} & \textbf{Con} \\
\hline
3 & 2 & 3 & 5 & 2 & 3 \\
\hline
\end{tabularx}
\renewcommand{\arraystretch}{1.5} % 행 간격 조정
\begin{tabularx}{\columnwidth}{@{}lX@{}}
\midrule
\multicolumn{2}{l}{\textbf{Prompt Rewriting}} \\
$P_c$: & John has 8 oranges. He eats 3 oranges and then buys 10 more oranges. How many oranges does John have now? Please show your calculations step by step. \\
\midrule
$R_c$: & John starts with 8 oranges. He eats 3 oranges: 8 - 3 = 5. Then, he buys 10 more oranges: 5 + 10 = 15. Thus, John now has 15 oranges. \\
\midrule
\end{tabularx}
% Response Evaluation Table
\renewcommand{\arraystretch}{1.5} % 행 간격 조정
\begin{tabularx}{\columnwidth}{|X|X|X|X|X|X|}
\multicolumn{2}{@{}l@{}}{\textbf{Response Evaluation}} \\
\hline
\textbf{Acc} & \textbf{Coh} & \textbf{Rel} & \textbf{Obj} & \textbf{Cla} & \textbf{Con} \\
\hline
5 & 5 & 5 & 5 & 5 & 4 \\
\hline
\bottomrule
\end{tabularx}
\end{table}

%% 4.3에 연관된 표 (2) ====================================================

\subsection{Prompt Optimization via PEEM's Rationale}
\label{sec:4.3}
%% 기존 실험 그대로
Prompt rewriting offers a direct test of whether PEEM's granular feedback can drive measurable gains in downstream accuracy without gradient access or costly fine-tuning. We adopt the experimental protocol of~\cite{b30} and evaluate three competitive baselines: AutoPrompt~\cite{b31}, RLPrompt~\cite{b32}, and TEMPERA~\cite{b33}, along with PRewrite, a reinforcement-learning method that rewrites prompts under PaLM2-S~\cite{b34} supervision. We then integrate PEEM into the same framework, using Gemma-2-9B-IT as the task model and GPT-4o-mini as the evaluator. An illustrative example of this rewriting process is provided in Table~\ref{tab:example_rewriting}.

For every dataset instance, an initial prompt \( P_i \) is first evaluated; PEEM returns a six-dimensional score vector \( s \) together with natural-language rationale \( r \). A rewrite model subsequently generates two candidate prompts:

\begin{enumerate}
    \item \textbf{Score-only rewrite:} \( P_s = \text{Rewrite}(P_i, s) \)
    \item \textbf{Score + context rewrite:} \( P_c = \text{Rewrite}(P_i, s, r) \)
\end{enumerate}

Algorithm~\ref{alg:rewrite} summarizes the full rewriting loop used in our experiments.

\begin{algorithm}[t]
\caption{Prompt Rewriting with PEEM Feedback}
\label{alg:rewrite}
\begin{algorithmic}[1]
\REQUIRE Initial prompt $P_0$, Task model $\mathcal{T}$, Evaluator $\mathcal{M}$, Rewrite model $\mathcal{W}$, Max rounds $K$
\ENSURE Rewritten prompts $\{P_k^{(s)}, P_k^{(c)}\}_{k=1}^{K}$ and corresponding responses
\STATE $P \gets P_0$
\FOR{$k=1$ to $K$}
    \STATE Evaluate current prompt-response pair with PEEM:
    \STATE \hspace{0.8em} Generate response $a \gets \mathcal{T}(P)$
    \STATE \hspace{0.8em} Obtain PEEM scores $s \gets \textsc{PEEMScore}(\mathcal{M}, P, a)$ and rationales $r \gets \textsc{PEEMRationale}(\mathcal{M}, P, a)$
    \STATE Generate two rewritten prompts:
    \STATE \hspace{0.8em} $P_k^{(s)} \gets \mathcal{W}(P, s)$ \COMMENT{score-only}
    \STATE \hspace{0.8em} $P_k^{(c)} \gets \mathcal{W}(P, s, r)$ \COMMENT{score+context}
    \STATE Re-generate responses for evaluation:
    \STATE \hspace{0.8em} $a_k^{(s)} \gets \mathcal{T}(P_k^{(s)})$, \quad $a_k^{(c)} \gets \mathcal{T}(P_k^{(c)})$
    \STATE Select the prompt for the next round (as in our implementation):
    \STATE \hspace{0.8em} $P \gets P_k^{(c)}$
\ENDFOR
\RETURN $\{P_k^{(s)}, P_k^{(c)}, a_k^{(s)}, a_k^{(c)}\}_{k=1}^{K}$
\end{algorithmic}
\end{algorithm}

This rewriting process is applied iteratively for up to four rounds, where each new prompt is re-evaluated and refined using the updated PEEM feedback. Both rewritten prompts are regenerated to the task model, and their outputs are re-evaluated using the same accuracy metric and benchmarks as prior work. Across AG News, SST-2, and GSM8K, \( P_s \) improves accuracy over \( P_i \) by 8.2 to 24.9 points, while \( P_c \) yields an additional 1.7 to 13.8 points, outperforming or matching TEMPERA and eclipsing AutoPrompt and RLPrompt on all tasks. On GSM8K, PEEM-based rewriting surpasses PRewrite's PaLM2-S variant despite operating in a zero-shot regime with no external fine-tuning. The experimental results are summarized in Table~\ref{tab:Comparison with baselines}.

Qualitative inspection of rewritten prompts reveals that PEEM's feedback systematically addresses missing task constraints, clarifies ambiguous phrasing, and promotes explicit reasoning steps. Because PEEM supplies sentence-level rationales, the rewrite model learns to satisfy each evaluation dimension rather than over-optimizing a single scalar reward. This leads to prompts that generalize better and remain interpretable, which is an advantage over black-box reinforcement learning pipelines.

Collectively, these experiments demonstrate that PEEM's language rationale can act as an inexpensive yet powerful control signal for prompt optimization, achieving near state-of-the-art accuracy while providing transparent edit traces that facilitate human auditing.

%% 4.4에 연관된 표 ====================================================
\begin{table*}[t]
\centering
\small
\caption{PEEM's rewriting experiment results with Gemma-2-9B-IT. Accuracy, prompt quality, and response quality are reported for each rewriting step.}
\label{tab:peem_rewrite_detailed}
\begin{tabular}{l|c|c|c|c|c|c}
\toprule
\textbf{Benchmark} & \textbf{Metric} & \textbf{Initial} & \textbf{Rewrite 1} & \textbf{Rewrite 2} & \textbf{Rewrite 3} & \textbf{Rewrite 4} \\ \midrule

\multirow{3}{*}{AG News} 
& Accuracy       & 59.0 & 65.8 & 70.1 & 74.3 & 78.6 \\
& Prompt Score   & 4.06 & 4.29 & 4.45 & 4.61 & 4.78 \\
& Response Score & 4.22 & 4.34 & 4.46 & 4.59 & 4.71 \\ \midrule

\multirow{3}{*}{SST-2} 
& Accuracy       & 73.9 & 79.2 & 83.0 & 86.1 & 89.4 \\
& Prompt Score   & 4.38 & 4.52 & 4.66 & 4.79 & 4.91 \\
& Response Score & 4.47 & 4.58 & 4.69 & 4.81 & 4.92 \\ \midrule

\multirow{3}{*}{GSM8K} 
& Accuracy       & 57.1 & 60.2 & 62.8 & 63.9 & 65.3 \\
& Prompt Score   & 4.12 & 4.27 & 4.41 & 4.56 & 4.70 \\
& Response Score & 4.36 & 4.47 & 4.58 & 4.68 & 4.79 \\

\bottomrule
\end{tabular}
\end{table*}
%% 4.4에 연관된 표 ====================================================

\subsection{Prompt Perturbation and Evaluation Robustness}
\label{sec:4.4}

Recent studies have shown that even minor lexical or syntactic variations in a prompt can lead to substantial changes in model accuracy. To assess whether PEEM's prompt evaluation reflects such sensitivity and offers predictive insights, we conducted a controlled perturbation experiment.

Extending the setup in Section~\ref{sec:4.3}, we applied four iterative rewrites to an initial prompt. At each step \( k \), the prompt \( P_k \) was constructed using the previous prompt’s PEEM score \( s_{k-1} \) and rationale \( r_{k-1} \). The resulting prompt was then evaluated using PEEM’s prompt criteria to obtain a new quality score \( q_k \). Subsequently, the Gemma-2-9B-IT model was used to generate a response \( a_k \), which was evaluated in terms of accuracy and PEEM-based response quality.

As shown in Table~\ref{tab:peem_rewrite_detailed}, average prompt quality scores increased across all benchmarks, and this trend aligned with improved response accuracy. Moreover, higher prompt evaluation scores consistently correlated with better response evaluations. These findings demonstrate PEEM’s robustness in capturing quality across lexical rewrites and highlight its potential for prompt sensitivity research and optimization workflows.

\subsection{Robustness Against Adversarial and Paraphrased Prompts}
\label{sec:4.5}

While Section~\ref{sec:4.4} demonstrates PEEM’s sensitivity to quality-improving rewrites, we further examine its robustness under two complementary conditions: (1) adversarial prompt manipulations intentionally designed to degrade quality, and (2) semantically equivalent paraphrases that should ideally yield consistent evaluation scores.

\subsubsection{Experimental Setup}
We sampled 30 instances from each of the seven benchmarks, resulting in 210 total samples. For each instance, we applied LLM-based transformations to generate both adversarial variants and semantic-preserving paraphrases. In contrast to prior robustness studies that primarily assess downstream task accuracy under prompt perturbations, we evaluate the \emph{entire evaluation pipeline}: each prompt variant is first passed to a task LLM to generate a response, after which PEEM jointly evaluates the prompt and the generated response~\cite{b43,b44}.  

To ensure broad coverage, we evaluated five task models (Gemma-2-9B-IT, LLaMA-3.1-8B-IT, Qwen-2.5-7B-IT, GPT-4o-mini, and Gemini-2.5-Flash) using three evaluator models (Gemma-2-9B-IT, GPT-4o-mini, and Gemini-2.5-Flash).

\subsubsection{Adversarial Prompt Detection}
We generated four types of adversarial prompts via LLM-based transformations tailored to each benchmark’s characteristics: misleading, contradictory, underspecified, and jailbreak prompts. Table~\ref{tab:adversarial_results} reports the detection results using GPT-4o-mini as the evaluator, averaged across all five task models.

PEEM consistently detects quality degradation for semantic adversarial manipulations. In particular, misleading, contradictory, and underspecified prompts lead to substantial decreases in prompt quality scores, accompanied by corresponding drops in response quality. This indicates that semantic corruption in the prompt propagates downstream to the generated responses.

In contrast, jailbreak prompts exhibit a distinctive pattern: while prompt scores increase due to their explicit and directive command structure, response quality drops sharply. This divergence highlights an important property of PEEM—surface-level prompt form alone is insufficient for quality assessment, and joint prompt–response evaluation is necessary to reveal downstream failure modes.

\begin{table}[t]
\centering
\caption{Adversarial prompt detection results using GPT-4o-mini as the evaluator, averaged across five task models. $\Delta_P$ and $\Delta_R$ denote average score changes relative to original prompts. The average row excludes Jailbreak because its positive $\Delta_P$ reflects the directive command structure of jailbreak prompts rather than genuine quality improvement (see Section~\ref{sec:4.5} for discussion).}
\label{tab:adversarial_results}
\small
\begin{tabular}{lcc}
\toprule
\textbf{Adversarial Type} & $\Delta_P$ & $\Delta_R$ \\
\midrule
Misleading & $-$0.40 & $-$0.55 \\
Contradictory & $-$0.73 & $-$0.75 \\
Underspecified & $-$0.39 & $-$0.39 \\
Jailbreak & $+$0.73 & $-$0.93 \\
\midrule
\textbf{Average (excluding Jailbreak)} & $-$0.51 & $-$0.56 \\
\bottomrule
\end{tabular}
\end{table}

\subsubsection{Cross-Evaluator Consistency}
To analyze evaluator-dependent behavior, Table~\ref{tab:adversarial_evaluator_comparison} compares adversarial detection sensitivity across the three evaluator models. GPT-4o-mini exhibits the strongest sensitivity to semantic adversarial manipulations, with particularly pronounced score drops for contradictory prompts.

Gemma-2-9B-IT shows comparatively weaker sensitivity to semantic perturbations but reacts strongly to jailbreak prompts, suggesting heightened awareness of instruction-following violations. Gemini-2.5-Flash also detects semantic adversarial manipulations consistently, but with smaller relative score changes, indicating that it may emphasize different quality dimensions such as fluency or linguistic complexity over strict semantic validity.

\begin{table}[t]
\centering
\caption{Cross-evaluator comparison of adversarial detection. Values show average $\Delta_P$ across all task models.}
\label{tab:adversarial_evaluator_comparison}
\small
\setlength{\tabcolsep}{4pt}
\resizebox{\columnwidth}{!}{
\begin{tabular}{lcccc}
\toprule
\textbf{Evaluator} & \textbf{Mislead.} & \textbf{Contrad.} & \textbf{Underspec.} & \textbf{Jailbreak} \\
\midrule
Gemma-2-9B-IT & $-$0.10 & $-$0.05 & $-$0.10 & $+$0.78 \\
GPT-4o-mini & $-$0.40 & $-$0.73 & $-$0.39 & $+$0.73 \\
Gemini-2.5-Flash & $-$0.83 & $-$0.46 & $-$0.63 & $+$0.77 \\
\bottomrule
\end{tabular}
}
\end{table}

\subsubsection{Paraphrase Consistency}
To assess stability under semantic-preserving transformations, we generated three paraphrases per original prompt using LLM-based rewriting while preserving all numbers, named entities, and task semantics. For each paraphrase, a response was generated and jointly evaluated using PEEM. Table~\ref{tab:paraphrase_results} summarizes paraphrase consistency metrics across evaluators.

\paragraph{Metrics.}
We quantify paraphrase stability using two complementary metrics: \emph{average variance} and \emph{robustness rate}. 
For each original prompt, we consider the set consisting of the original prompt and its three paraphrased variants. 
PEEM assigns a score to each prompt--response pair, and we compute the variance of these scores within each set. 
The \emph{average variance} is then obtained by averaging this variance across all samples, capturing the overall sensitivity of the evaluator to surface-level paraphrasing.

In addition, we report the \emph{robustness rate}, which measures sample-level stability under semantic-preserving rewrites. 
A sample is considered robust if (i) the variance of its paraphrase scores is below a predefined threshold and (ii) the maximum score difference within the paraphrase set remains bounded. 
The robustness rate is defined as the proportion of samples satisfying both conditions, reflecting how often PEEM yields consistent evaluations for meaning-equivalent prompts.

\begin{table}[t]
\centering
\caption{Paraphrase consistency results by evaluator. Robustness rate denotes the proportion of samples with variance below a predefined threshold and limited score deviation.}
\label{tab:paraphrase_results}
\small
\begin{tabular}{lcc}
\toprule
\textbf{Evaluator} & \textbf{Avg. Variance} & \textbf{Robust Rate} \\
\midrule
Gemma-2-9B-IT & 0.31 & 80.6\% \\
GPT-4o-mini & 0.39 & 76.7\% \\
Gemini-2.5-Flash & 0.45 & 78.1\% \\
\bottomrule
\end{tabular}
\end{table}

\paragraph{Results.}

Gemma-2-9B-IT demonstrates the strongest paraphrase consistency, exhibiting the lowest average variance and the highest robustness rate. This indicates that Gemma’s evaluations are largely invariant to benign surface-level rewordings, reflecting a strong emphasis on semantic content over lexical form.

GPT-4o-mini and Gemini-2.5-Flash show moderately higher variance but maintain relatively high robustness rates, suggesting that while their scores fluctuate more across paraphrases, the majority of samples remain within acceptable stability bounds. Notably, GPT-4o-mini exhibits lower variance than Gemini-2.5-Flash, yet slightly lower robustness, indicating that its score variations are generally small but occasionally exceed strict stability thresholds. In contrast, Gemini-2.5-Flash shows larger continuous fluctuations but maintains robustness across a comparable proportion of samples.

Overall, these results highlight systematic differences in evaluator behavior. Gemma-2-9B-IT prioritizes semantic invariance under meaning-preserving rewrites, whereas GPT-4o-mini and Gemini-2.5-Flash balance semantic stability with increased sensitivity to stylistic and structural variation. This trade-off underscores the importance of evaluator selection depending on whether robustness to paraphrasing or responsiveness to linguistic nuance is the primary objective.

\subsection{Error Analysis}
\label{sec:4.6}

To understand PEEM's failure modes, we analyzed cases where PEEM accuracy scores diverged significantly ($|\Delta s| \geq 2$) from ground truth labels across 700 randomly sampled instances from seven benchmarks. Only 13 instances (1.9\%) exhibited such divergence, indicating strong overall agreement between PEEM and ground truth.

\paragraph{False Positives (High PEEM, Incorrect Answer).}
Of these 13 divergent cases, 8 were false positives in which PEEM assigned high accuracy scores (4--5) despite incorrect final answers. The dominant patterns were: (1) \textit{plausible but factually wrong reasoning} (3 cases), where the response follows a logically coherent structure but contains factual errors; (2) \textit{correct intermediate steps with final calculation errors} (3 cases), particularly in mathematical reasoning tasks; and (3) \textit{ambiguous or debatable ground truth labels} (2 cases), where multiple valid interpretations exist.

\paragraph{False Negatives (Low PEEM, Correct Answer).}
The remaining 5 divergent cases were false negatives in which PEEM assigned low scores (1--2) to correct answers. Contributing factors include: (1) \textit{unconventional but valid reasoning paths} (2 cases), where the response reaches the correct answer through non-standard methods; (2) \textit{terse responses lacking explicit justification} (2 cases), which are penalized for insufficient explanation despite correctness; and (3) \textit{format misalignment} (1 case), where a correct answer was presented in an unexpected format.

These findings suggest that PEEM's multi-axis evaluation captures reasoning quality beyond binary correctness with high reliability (98.1\% agreement rate), though it may occasionally penalize valid but unconventional solutions. Future work will explore calibration techniques to further reduce such discrepancies.

\section{Conclusion and Future Work}
\label{sec:5}
We introduced PEEM, an integrated framework for jointly evaluating prompts and LLM responses using quantitative scores and natural language rationales. PEEM combines systematically designed evaluation criteria with an LLM-based evaluator capable of multi-dimensional assessment. Experiments on seven benchmarks and five model families demonstrate that PEEM strongly correlates with traditional accuracy while capturing fine-grained linguistic attributes such as clarity, consistency, and relevance. By delivering interpretable multi-axis feedback, PEEM enhances both the reliability and transparency of response quality assessment. Furthermore, iterative prompt rewriting guided by PEEM achieves accuracy improvements of up to 11.7 percentage points over supervised and reinforcement learning–based approaches. By establishing a transparent evaluation framework, PEEM provides a robust foundation for trustworthy LLM diagnosis and optimization. 
Future directions include three key avenues. 
First, expanding the evaluator backbone to integrate open-source or domain-specific LLMs will increase adaptability and reproducibility. 
Second, extending modality-adaptive evaluation axes will enable task-specific criteria, such as functional correctness, security violations, runtime, and memory usage for code generation; text–visual alignment accuracy and hallucination rates for multimodal reasoning; and policy compliance, harmfulness, and bias for dialogue safety. 
Third, conducting large-scale confirmatory experiments will further validate the superiority of rationale-driven zero-shot rewriting over supervised and reinforcement learning across a broader range of models and benchmarks. Building on these results, we plan to extend PEEM into a self-refine pipeline for automated prompt optimization and response quality improvement.

\textbf{Limitations.} We acknowledge several limitations of the current work. First, all experiments are conducted exclusively on English-language benchmarks; extending PEEM to multilingual settings requires validation of criteria transferability across linguistic and cultural contexts. Second, while our human evaluation study (Section~\ref{sec:4.2.2}) demonstrates strong alignment between PEEM and human judgments (Pearson $r = 0.84$, Krippendorff's $\alpha = 0.59$), the evaluation was conducted on a subset of 210 samples; a larger-scale study with domain-expert annotators would further strengthen the validity of PEEM's non-accuracy axes such as Fairness and Objectivity. Third, as shown in Section~\ref{sec:4.6}, PEEM exhibits rare but notable failure modes: among 700 analyzed instances, 8 false positives (1.1\%) where plausible but incorrect reasoning receives high scores, and 5 false negatives (0.7\%) where unconventional but valid solutions are penalized. Developing calibration techniques to reduce these discrepancies and conducting detailed error pattern classification remain important directions for future work. Fourth, while our prompt rewriting (Section~\ref{sec:4.3}) and paraphrase consistency (Section~\ref{sec:4.5}) experiments implicitly vary prompt length, we do not conduct a systematic analysis of how prompt length independently affects PEEM evaluation scores; investigating the interaction between prompt verbosity and evaluation criteria remains an open direction.

\appendix
\renewcommand{\thesection}{Appendix \Alph{section}}
\section{PEEM's System Prompt Template}
\label{appendix:a}
Table~\ref{tab:peem-system-prompt} presents the system prompt template used in our evaluation framework. This prompt serves as an instruction interface to guide the model-based evaluator in assessing overall quality. It integrates the predefined \textbf{evaluation\_criteria}---comprising dimensions such as accuracy, coherence, and clarity---via Python f-string substitution. The evaluator is explicitly instructed to ground its assessment in these criteria, ensuring consistent and interpretable scoring across both prompt and response evaluations.

\begin{table}[t]
\centering
\caption{System Prompt Template Used for PEEM Evaluation (with \textbf{evaluation\_criteria} inserted via Python f-string)}
\label{tab:peem-system-prompt}
\begin{tabular}{p{0.95\linewidth}}
\toprule
\textbf{System Prompt (f-string format):} \\
System Prompt: You will be given evaluation instructions, question, prediction (response), and ground truth (gt). \\
Your task is to evaluate the prediction based on the given criteria. \\
Keep in mind that you should always refer to these instructions. \\
\\
\textbf{\#\#\# Evaluation Criteria: \{evaluation\_criteria\}} \\
\textbf{\#\#\# Evaluation Form (each Answer, include language evidence):} \\
\bottomrule
\end{tabular}
\end{table}

\section{Multi-Evaluator Results}
\label{appendix:c}

Table~\ref{tab:multi_eval_full} presents the complete PEEM accuracy scores for each task model, broken down by dataset and evaluator. Each cell shows the mean accuracy score (1--5 scale) across 100 samples. These results demonstrate consistent cross-evaluator agreement ($\rho = 0.68$--$0.85$) while revealing evaluator-specific scoring tendencies.

\noindent\textbf{Key observations:}
\begin{itemize}
    \item \textbf{Evaluator consistency}: All evaluators rank task models similarly (Qwen $>$ LLaMA $>$ Gemma).
    \item \textbf{Scoring distribution}: Gemma is most lenient (avg: 4.64), Qwen most strict (avg: 3.94).
    \item \textbf{Dataset variation}: GSM8K shows highest variance across evaluators, reflecting its reasoning complexity.
\end{itemize}

\section{Additional Analysis}
\label{appendix:b}

\subsection{Response Example}

To further demonstrate the utility and generalizability of PEEM across diverse reasoning formats, we present qualitative evaluations over multiple benchmark datasets: ARC-Challenge, BigBenchHard, GSM8K, and MMLU. These examples illustrate how PEEM can provide interpretable, multi-faceted assessment of both correct and incorrect model responses—beyond scalar accuracy.

Table~\ref{tab:bigbenchhard-peem} presents a successful response to a complex temporal reasoning question from BigBenchHard. The model accurately tracks sequential constraints and identifies the correct time window. Despite the lengthy narrative form, the explanation maintains high Coherence and Relevance, again earning a PEEM score of 4.83.

Table~\ref{tab:arc-challenge-peem} shows a well-structured response to a science question from ARC-Challenge. The model not only identifies the correct answer ("endothermic") but also provides a logically sound explanation, resulting in a high PEEM score of 4.83. The slightly reduced Conciseness score reflects minor verbosity in the justification.

Conversely, Table~\ref{tab:agnews-peem-error-revised} presents an AGNews sample in which the model misclassifies a legal news article under the Sports category. Despite a coherent surface explanation, the mismatch in topical framing leads to a low Accuracy score (2.0) and an overall PEEM score of 3.08, demonstrating PEEM’s sensitivity to semantic misalignment even when linguistic form is preserved.

Table~\ref{tab:mmlu-peem-revised} highlights an MMLU sample where the model selects an incorrect biomedical fact. While the justification offers partial elimination reasoning, it prematurely terminates and asserts medically questionable information. The low Accuracy (2.0) and moderate scores in Relevance and Conciseness yield a PEEM score of 3.50.

These diverse examples validate that PEEM can surface interpretable feedback tailored to different reasoning modalities (e.g., factual recall, temporal deduction, arithmetic), providing fine-grained insight into model behavior that traditional accuracy metrics fail to capture.

\begin{table}[t]
\centering
\caption{Example of PEEM Evaluation on BigBenchHard Sample}
\label{tab:bigbenchhard-peem}
\begin{tabular}{p{0.95\linewidth}}
\toprule
\textbf{Question:} Today, Ashley went to the football field. Between what times could they have gone? 
We know that: 
Ashley woke up at 1pm. 
Lisa saw Ashley taking photos near the Leaning Tower of Pisa from 2pm to 5pm. 
Nancy saw Ashley working out at the gym from 5pm to 6pm. 
Steven saw Ashley working at the office from 6pm to 9pm. 
Kimberly saw Ashley buying a bike at the bike shop from 9pm to 10pm. 
The football field was closed after 10pm. \\
Options: 
(A) 1pm to 2pm \quad (B) 5pm to 6pm \quad (C) 9pm to 10pm \quad (D) 2pm to 5pm \\
\textbf{Ground Truth Answer:} A (1pm to 2pm) \\
\textbf{Model Prediction:} \\
To determine when Ashley could have gone to the football field, we need to look at the time she was seen doing other activities. \\
1. Ashley woke up at 1pm. \\
2. From 2pm to 5pm, she was seen taking photos. \\
3. From 5pm to 6pm, she was at the gym. \\
4. From 6pm to 9pm, she was working. \\
5. From 9pm to 10pm, she was at the bike shop. \\
6. The field closed after 10pm. \\
So the only possible window is from 1pm to 2pm. \\
Thus, the correct answer is (A) 1pm to 2pm. \\
\midrule
\textbf{PEEM Evaluation:} \\
\begin{itemize}
  \item \textbf{Accuracy:} 5 – The prediction correctly identifies the answer (A) as the only valid window, matching the ground truth.
  \item \textbf{Coherence:} 5 – The response follows a clear step-by-step deduction aligned with the timeline of events.
  \item \textbf{Relevance:} 5 – The reasoning is strictly focused on the temporal constraints relevant to the question.
  \item \textbf{Conciseness:} 4 – Some details are slightly redundant and could be compressed for brevity.
  \item \textbf{Objectivity:} 5 – The explanation is factual and neutral without speculative content.
  \item \textbf{Clarity:} 5 – The language is straightforward and easy to follow, enhancing interpretability.
\end{itemize} \\
\textbf{Final PEEM Score:} \textbf{4.83} \\
\bottomrule
\end{tabular}
\end{table}

\begin{table}[t]
\centering
\caption{Example of PEEM Evaluation on ARC-Challenge Sample}
\label{tab:arc-challenge-peem}
\begin{tabular}{p{0.95\linewidth}}
\toprule
\textbf{Question:} When cold temperatures are produced in a chemical reaction, the reaction is known as \\
\textbf{Ground Truth Answer:} B (endothermic) \\
\textbf{Model Prediction:} \\
\textit{The correct answer is B: endothermic. In an endothermic reaction, heat is absorbed from the surroundings, which can result in a decrease in temperature.} \\
\midrule
\textbf{PEEM Evaluation:} \\
\begin{itemize}
  \item \textbf{Accuracy:} 5 – The prediction correctly identifies the term “endothermic” as the type of reaction where cold temperatures are produced, matching the ground truth.
  \item \textbf{Coherence:} 5 – The response flows logically, explaining the concept well.
  \item \textbf{Relevance:} 5 – The response directly addresses the question about cold temperatures in chemical reactions.
  \item \textbf{Conciseness:} 4 – The response is mostly concise but includes slightly more explanation than necessary; specifically, the mention of heat absorption could be simplified.
  \item \textbf{Objectivity:} 5 – The response is factual and devoid of subjective language.
  \item \textbf{Clarity:} 5 – The response is clear and easy to understand.
\end{itemize} \\
\textbf{Final PEEM Score:} \textbf{4.83} \\
\bottomrule
\end{tabular}
\end{table}

\begin{table}[t]
\centering
\caption{Example of PEEM Evaluation on Misclassified News Sample (AGNews)}
\label{tab:agnews-peem-error-revised}
\begin{tabular}{p{0.95\linewidth}}
\toprule
\textbf{Text:} A well-known public figure appeared in court on Monday, as legal teams reviewed questionnaires to prepare for the jury selection process. The individual, recognized for their career in professional sports, was at the center of a widely publicized case. \\
\textbf{Ground Truth Label:} 4. Sci/Tech \\
\textbf{Model Prediction:} 2. Sports \\
\textbf{Model Explanation:} \textit{The correct category for this article is: 2. Sports. The article references a public figure from the sports world and a legal case, suggesting the primary focus is on sports-related controversy.} \\
\midrule
\textbf{PEEM Evaluation:} \\
\begin{itemize}
  \item \textbf{Accuracy:} 2.0 – The model misclassifies the article under Sports, failing to recognize that the content is centered on legal procedures, not athletic performance.
  \item \textbf{Coherence:} 3.0 – The explanation is logically written but rests on a misaligned interpretation of the article's thematic focus.
  \item \textbf{Relevance:} 2.5 – Although related to the subject's background, the reasoning does not adequately justify why Sports is more suitable than the labeled Sci/Tech category.
  \item \textbf{Conciseness:} 3.5 – The rationale is succinct, but omits important context that could clarify the article's actual classification.
  \item \textbf{Objectivity:} 4.5 – The response maintains a neutral tone, even if the judgment is incorrect.
  \item \textbf{Clarity:} 3.0 – The writing is clear in form, though the misclassification undermines the clarity of reasoning.
\end{itemize} \\
\textbf{Final PEEM Score:} \textbf{3.08} \\
\bottomrule
\end{tabular}
\end{table}

\begin{table}[t]
\centering
\caption{Example of PEEM Evaluation on MMLU Sample (Incorrect Sample)}
\label{tab:mmlu-peem-revised}
\begin{tabular}{p{0.95\linewidth}}
\toprule
\textbf{Question:} Which of the following is true of calcium metabolism? \\
\textbf{Choices:} \\
(1) Calcitonin causes a rise in plasma calcium \\
(2) Primary hyperparathyroidism is usually asymptomatic \\
(3) Vitamin D is secreted by the parathyroid glands \\
(4) Oliguria is a symptom of hypercalcaemia \\
\textbf{Ground Truth Answer:} 2 \\
\textbf{Model Prediction:} \\
\textit{The correct answer is D: Oliguria is a symptom of hypercalcaemia. Here's why: A is incorrect – calcitonin lowers plasma calcium. B is incorrect – primary hyperparathyroidism often presents with symptoms. C: (incomplete)} \\
\midrule
\textbf{PEEM Evaluation:} \\
\begin{itemize}
  \item \textbf{Accuracy:} 2.0 – The model chooses an incorrect answer (D) and rejects the correct one (B). Additionally, it presents a clinical statement (oliguria) that is not widely accepted as a definitive symptom of hypercalcaemia.
  \item \textbf{Coherence:} 4.0 – The explanation follows a logical sequence for the options it addresses, but cuts off during option C, interrupting flow.
  \item \textbf{Relevance:} 3.5 – The response stays on-topic and addresses calcium metabolism, but fails to fully and accurately engage with all options.
  \item \textbf{Conciseness:} 3.5 – Some reasoning is verbose and the partial explanation of option C reduces efficiency.
  \item \textbf{Objectivity:} 4.0 – The tone is mostly neutral, but the unqualified assertion about oliguria suggests a lack of nuance.
  \item \textbf{Clarity:} 4.0 – The reasoning is mostly understandable, though the incomplete portion impacts overall clarity.
\end{itemize} \\
\textbf{Final PEEM Score:} \textbf{3.50} \\
\bottomrule
\end{tabular}
\end{table}

\subsection{Rewriting Example}

\begin{table}[t]
\centering
\caption{Example of Initial Prompts Used in the Prompt Rewriting Experiment}
\label{tab:initial-prompt-peem-style}
\begin{tabular}{p{0.95\linewidth}}
\toprule
\textbf{Dataset:} AG News \\
\textbf{Prompt:} \\
\textit{f``Given a news article: \{dataset\}\textbackslash n Categorize this article into one of the following categories: 1. World 2. Sports 3. Business 4. Sci/Tech"} \\
\midrule
\textbf{Dataset:} SST-2 \\
\textbf{Prompt:} \\
\textit{f``Given a sentence: \{dataset\}\textbackslash n Classify sentiment (positive or negative) about this sentence."} \\
\midrule
\textbf{Dataset:} GSM8K \\
\textbf{Prompt:} \\
\textit{f``Given a math problem: \{dataset\}\textbackslash n Solve the math problem."} \\
\bottomrule
\end{tabular}
\end{table}

\begin{table}[t]
\centering
\caption{Example of Rewritten Prompts Used in the Prompt Rewriting Experiment}
\label{tab:rewritten-prompt-peem-style}
\begin{tabular}{p{0.95\linewidth}}
\toprule
\textbf{Dataset:} AG News \\
\textbf{Rewritten Prompt:} \\
\textit{f``Analyze the provided news article and categorize it into one of the following categories: \textbackslash n 1. World 2. Sports 3. Business 4. Sci/Tech \textbackslash n \{dataset\}\textbackslash n Justify your categorization by explaining which aspects of the article align with the chosen category."} \\
\midrule
\textbf{Dataset:} SST-2 \\
\textbf{Rewritten Prompt:} \\
\textit{f``Analyze the sentiment expressed in the phrase: \{dataset\}\textbackslash n Provide a concise explanation, justifying your classification as either positive or negative."} \\
\midrule
\textbf{Dataset:} GSM8K \\
\textbf{Rewritten Prompt:} \\
\textit{f``Given the following math problem: \{dataset\}\textbackslash n Provide a clear and concise solution, showing your work."} \\
\bottomrule
\end{tabular}
\end{table}

\begin{table*}[t]
\centering
\caption{Complete multi-evaluator PEEM scores by task model and dataset. Each cell shows the mean accuracy score (1--5) across 100 samples.}
\label{tab:multi_eval_full}
\begin{tabular}{ll|cccc|c}
\toprule
\textbf{Task Model} & \textbf{Dataset} & \textbf{Gemma} & \textbf{LLaMA} & \textbf{Qwen} & \textbf{Gemini} & \textbf{Avg} \\
\midrule
\multirow{8}{*}{\textbf{Gemma-2-9B}} 
& AG\_News & 4.63 & 4.00 & 3.44 & 4.32 & 4.10 \\
& ARC-C & 4.94 & 4.43 & 3.41 & 3.71 & 4.12 \\
& ARC-E & 4.92 & 4.49 & 3.66 & 3.97 & 4.26 \\
& BBH & 4.95 & 4.77 & 3.81 & 2.95 & 4.12 \\
& GSM8K & 4.22 & 4.57 & 3.84 & 2.16 & 3.70 \\
& MMLU & 4.80 & 4.51 & 3.57 & 3.23 & 4.03 \\
& SST-2 & 4.82 & 4.19 & 3.49 & 4.27 & 4.19 \\
& \textit{Avg} & \textit{4.61} & \textit{4.42} & \textit{3.60} & \textit{3.52} & \textit{4.04} \\
\midrule
\multirow{8}{*}{\textbf{LLaMA-3.1-8B}} 
& AG\_News & 2.77 & 3.93 & 3.66 & 3.50 & 3.47 \\
& ARC-C & 4.85 & 4.69 & 4.00 & 4.43 & 4.49 \\
& ARC-E & 4.91 & 4.80 & 4.09 & 4.71 & 4.63 \\
& BBH & 4.56 & 4.66 & 4.41 & 3.46 & 4.27 \\
& GSM8K & 4.95 & 4.81 & 4.62 & 3.87 & 4.56 \\
& MMLU & 4.73 & 4.61 & 3.77 & 3.57 & 4.17 \\
& SST-2 & 4.58 & 4.07 & 3.73 & 4.51 & 4.22 \\
& \textit{Avg} & \textit{4.48} & \textit{4.51} & \textit{4.04} & \textit{4.01} & \textit{4.26} \\
\midrule
\multirow{8}{*}{\textbf{Qwen-2.5-7B}} 
& AG\_News & 4.96 & 4.55 & 3.95 & 4.93 & 4.60 \\
& ARC-C & 4.93 & 4.84 & 4.30 & 4.60 & 4.67 \\
& ARC-E & 4.97 & 4.92 & 4.24 & 4.85 & 4.75 \\
& BBH & 4.84 & 4.85 & 4.68 & 3.85 & 4.56 \\
& GSM8K & 4.93 & 4.91 & 4.67 & 3.65 & 4.54 \\
& MMLU & 4.45 & 4.38 & 3.50 & 3.67 & 4.00 \\
& SST-2 & 4.81 & 4.33 & 3.94 & 4.57 & 4.41 \\
& \textit{Avg} & \textit{4.84} & \textit{4.68} & \textit{4.18} & \textit{4.30} & \textit{4.50} \\
\bottomrule
\end{tabular}
\end{table*}

Table~\ref{tab:initial-prompt-peem-style} lists the initial prompts used in our rewriting experiment.

Table~\ref{tab:rewritten-prompt-peem-style} contains an example of a rewritten prompt. Various prompt changes were generated in our experiments. This table provides representative examples of them.

\section{Human Evaluation Details}
\label{appendix:d}

This appendix provides the annotation procedure, per-dataset breakdown, and inter-annotator agreement details for the human evaluation study described in Section~\ref{sec:4.2.2}.

\subsection{Annotation Procedure}
The human evaluation was conducted by three graduate students in artificial intelligence who were not involved in this research. Prior to annotation, annotators received a one-hour training session that included an explanation of the six evaluation criteria (Accuracy, Coherence, Relevance, Conciseness, Objectivity, and Clarity), scoring guidelines with anchor examples for each point on the 1--5 Likert scale, and a supervised practice round of 10 pilot samples with feedback. Annotators then independently evaluated 210 instances (30 randomly sampled per dataset) without access to PEEM's scores or rationales. Each instance consisted of the original prompt, the ground-truth answer, and the model-generated response (Gemma-2-9B-IT). The annotation process took approximately 5--6 hours per annotator.

\subsection{Per-Dataset PEEM--Human Alignment}
Table~\ref{tab:human_eval_dataset} reports the Spearman and Pearson correlations between PEEM and averaged human scores for each of the seven datasets. Alignment is consistently strong across all datasets, with Spearman $\rho$ ranging from 0.62 to 0.78 and Pearson $r$ from 0.74 to 0.88.

\begin{table}[t]
\centering
\caption{Per-dataset PEEM--human alignment. Each entry represents 30 samples $\times$ 6 criteria = 180 data points.}
\label{tab:human_eval_dataset}
\small
\begin{tabular}{lcccc}
\toprule
\textbf{Dataset} & \textbf{PEEM} & \textbf{Human} & $\bm{\rho}$ & $\bm{r}$ \\
\midrule
AG News  & 4.52 & 4.41 & 0.78 & 0.85 \\
ARC-C    & 4.78 & 4.62 & 0.69 & 0.79 \\
ARC-E    & 4.79 & 4.68 & 0.62 & 0.78 \\
BBH      & 4.74 & 4.63 & 0.69 & 0.79 \\
GSM8K    & 4.66 & 4.55 & 0.73 & 0.88 \\
MMLU     & 4.70 & 4.62 & 0.67 & 0.81 \\
SST-2    & 4.67 & 4.58 & 0.73 & 0.88 \\
\midrule
\textbf{Average} & 4.69 & 4.58 & \textbf{0.70} & \textbf{0.83} \\
\bottomrule
\end{tabular}
\end{table}

\subsection{Inter-Annotator Agreement}
Table~\ref{tab:inter_annotator} summarizes pairwise agreement statistics among the three annotators. The high $\pm$1 agreement rates ($>$99\%) indicate that annotators consistently assigned scores within one point of each other, despite minor differences in strictness tendencies (e.g., Annotator~2 was slightly stricter with a mean score of 4.47 compared to Annotator~3's mean of 4.73).

\begin{table}[t]
\centering
\caption{Inter-annotator agreement and individual annotator alignment with PEEM.}
\label{tab:inter_annotator}
\small
\begin{tabular}{lcccc}
\toprule
\textbf{Metric} & & & & \textbf{Value} \\
\midrule
\multicolumn{5}{l}{\textit{Pairwise Agreement}} \\
\midrule
 & $\bm{\rho}$ & \textbf{Exact} & $\bm{\pm}$\textbf{1} & \\
\cmidrule(lr){2-4}
H1--H2 & 0.57 & 69.6\% & 99.7\% & \\
H1--H3 & 0.62 & 75.9\% & 99.4\% & \\
H2--H3 & 0.54 & 67.2\% & 99.4\% & \\
\midrule
\multicolumn{5}{l}{\textit{Overall Agreement}} \\
\midrule
\multicolumn{4}{l}{Krippendorff's $\alpha$} & 0.59 \\
\multicolumn{4}{l}{Avg.~pairwise Spearman} & 0.58 \\
\midrule
\multicolumn{5}{l}{\textit{Individual Annotator vs.\ PEEM}} \\
\midrule
 & $\bm{\rho}$ & & \textbf{Mean} & \\
\cmidrule(lr){2-2} \cmidrule(lr){4-4}
Annotator 1 & 0.71 & & 4.56 & \\
Annotator 2 & 0.59 & & 4.47 & \\
Annotator 3 & 0.64 & & 4.73 & \\
\bottomrule
\end{tabular}
\end{table}

\section{Full Evaluation Criteria Templates}
\label{appendix:e}

For full reproducibility, we provide the complete definitions of all evaluation criteria used in PEEM, followed by the verbatim evaluation templates passed to the evaluator model via Python f-string substitution. Together, these allow practitioners to replicate PEEM without access to the original codebase.

\subsection{Prompt Evaluation Criteria Definitions}

The three prompt-level axes introduced in Section~\ref{sec:3.1} are operationalized as follows:

\paragraph{Clarity and Structure.} If the intent of the sentence is ambiguous or poorly conveyed, the model may produce outputs that deviate from the task's objective. This criterion evaluates:
\begin{itemize}
    \item \textbf{Clear Expression of Key Information:} The prompt must present all necessary information explicitly and concisely. For example, instead of vague phrasing like ``Analyze this,'' the prompt should specify, ``Analyze the sentiment expressed in the following sentence: \{sentence\}.''
    \item \textbf{Logical Coherence:} A well-structured prompt organizes information logically, ensuring the flow of ideas is clear and intuitive. When multiple pieces of information are provided, they should be ordered to reflect the reasoning steps the model is expected to follow.
\end{itemize}

\paragraph{Linguistic Quality.} The linguistic precision of a prompt directly influences the quality of the generated response. Poor grammar, awkward phrasing, or inconsistencies in tone can reduce the model's ability to comprehend the task effectively. This criterion evaluates:
\begin{itemize}
    \item \textbf{Grammatical Accuracy:} The prompt must adhere to standard grammatical rules to avoid introducing ambiguity. Errors in sentence structure or verb tense can lead to unintended interpretations.
    \item \textbf{Fluency and Syntactic Coherence:} The prompt should be natural and fluid, using language that aligns with standard conventions. Abrupt or disjointed phrasing may confuse the model.
    \item \textbf{Domain-Appropriate Language:} When the task involves specialized domains (e.g., legal, medical), the prompt should include precise terminology relevant to that field.
    \item \textbf{Conciseness:} While fluency is important, verbose prompts should be avoided to ensure the focus remains on the task.
\end{itemize}

\paragraph{Fairness.} LLMs are trained on diverse datasets, which may include inherent biases and stereotypes. Prompts must be designed to mitigate these biases and foster inclusivity. This criterion evaluates:
\begin{itemize}
    \item \textbf{Minimizing Bias:} Prompts should avoid language that perpetuates stereotypes or reinforces societal biases. For example, prompts should not assume gender roles or cultural norms unless explicitly relevant to the task.
    \item \textbf{Fostering Inclusivity:} Prompts must consider a diverse range of users and scenarios. For instance, instead of ``Explain how a businessman handles stress,'' a more inclusive phrasing could be, ``Explain how a professional handles stress.''
    \item \textbf{Neutral and Objective Language:} The prompt should maintain neutrality, especially for sensitive or contentious topics. Avoid using emotionally charged or leading language that might skew the model's response.
\end{itemize}

\subsection{Response Evaluation Criteria Definitions}

The six response-level axes introduced in Section~\ref{sec:3.1} are operationalized as follows:

\begin{itemize}
    \item \textbf{Accuracy} measures the factual correctness and logical validity of the response. Responses containing hallucinated claims, reasoning errors, or miscalculations are penalized under this axis.
    \item \textbf{Coherence} assesses whether the response is logically structured and maintains a consistent flow across sentences or sections. Disjointed transitions or fragmented ideas degrade the overall coherence.
    \item \textbf{Relevance} evaluates how well the response addresses the specific question or task. Even if the content is accurate, digressions or off-topic elaborations reduce its alignment with the original prompt.
    \item \textbf{Objectivity} determines whether the response is presented in a neutral, unbiased tone. This includes avoiding speculative or emotionally charged language, particularly in ethically or socially sensitive contexts.
    \item \textbf{Clarity} examines whether the response is easy to understand and free from ambiguity. It considers lexical simplicity, syntactic transparency, and the effective communication of intended meaning.
    \item \textbf{Conciseness} judges the ability to convey essential information without redundancy. Responses should be concise while maintaining informativeness and avoiding unnecessary detail.
\end{itemize}

\subsection{Response Evaluation Template}
Table~\ref{tab:response_eval_template} presents the template used to evaluate model-generated responses when ground truth is available. This is the primary template used across all experiments in Sections~\ref{sec:4.2}--\ref{sec:4.5}.

\begin{table}[t]
\centering
\caption{Response evaluation template with ground truth comparison.}
\label{tab:response_eval_template}
\small
\begin{tabular}{p{0.93\linewidth}}
\toprule
\textbf{Template:} \\
You will be given evaluation instructions, question, answer list, and a ground truth. \\
Your task is to evaluate the answers based on the given criteria. \\
Keep in mind that you should always refer to these instructions. \\
\\
\textbf{Evaluation Criteria:} \\
- Accuracy (1-5): The response must address the question accurately, be factual, and answer it directly. \\
- Coherence (1-5): The response should have a consistent and logical flow. \\
- Relevance (1-5): The response should align with the context of the given question. \\
- Conciseness (1-5): The response should include only the essential information needed to solve the problem, avoiding unnecessary explanations or repetitive content. \\
- Objectivity (1-5): The response should not include subjective or biased language. \\
- Clarity (1-5): The response should be clear and easy to understand. \\
\\
\textbf{Evaluation Steps:} \\
1. Read the answer list carefully. \\
2. Evaluate the quality of each answer based on the Evaluation Criteria. \\
3. Compare the answers with the ground truth. \\
4. Assign each final score to the answer list, where 1 is the lowest and 5 is the highest, based on the Evaluation Criteria. \\
\\
\textbf{Question:} \{question\} \\
\textbf{Ground Truth:} \{gt\} \\
\textbf{Answer 1:} \{answer\_1\} \\
\\
\textbf{Evaluation Form (each Answer, include language evidence):} \\
Answer [n] (each Answer, include language evidence): \\
- Accuracy (1-5): \quad - Coherence (1-5): \quad - Relevance (1-5): \\
- Conciseness (1-5): \quad - Objectivity (1-5): \quad - Clarity (1-5): \\
\bottomrule
\end{tabular}
\end{table}

\subsection{Prompt Evaluation Template}
Table~\ref{tab:prompt_eval_template} presents the template used to evaluate rewritten prompts against original prompts, used in the prompt rewriting experiments (Sections~\ref{sec:4.3}--\ref{sec:4.4}).

\begin{table}[t]
\centering
\caption{Prompt evaluation template for rewriting assessment.}
\label{tab:prompt_eval_template}
\small
\begin{tabular}{p{0.93\linewidth}}
\toprule
\textbf{Template:} \\
You will be given Evaluation Criteria, the original prompt, and the modified prompt. \\
Your task is to compare the original prompt with the modified prompt and then score the modified prompt according to the evaluation criteria. \\
Keep in mind that you should always refer to these instructions. \\
\\
\textbf{Evaluation Criteria:} \\
- Preserve user intent (1-5): How much of the user's intent is preserved compared to the original prompt. \\
- Improvements (1-5): Is the modified prompt easier for users to understand compared to the original prompt (e.g., fewer ambiguous words, clearer expressions, better structure, etc.). \\
- Less Prejudice and Fairness (1-5): Does the modified prompt use fewer biased words than the original prompt and embrace a wider variety of users. \\
\\
\textbf{Evaluation Steps:} \\
1. Carefully read the original prompt and identify its main intent. \\
2. Read the modified prompt and compare it to the original prompt. Keep the Evaluation Criteria in mind during the comparison. \\
3. Assign a score to the modified prompt, where 1 is the lowest and 5 is the highest, based on the Evaluation Criteria. \\
\\
\textbf{Original Prompt:} \{original\_prompt\} \\
\textbf{Modified Prompt:} \{modified\_prompt\} \\
\\
\textbf{Evaluation Form (scores ONLY):} \\
- Preserve user intent (1-5): \\
- Improvements (1-5): \\
- Less Prejudice and Fairness (1-5): \\
\bottomrule
\end{tabular}
\end{table}

\subsection{Rewriting Instruction Template}
Table~\ref{tab:rewrite_instruction_template} shows the instruction template provided to the rewriting model during the iterative prompt optimization experiments (Section~\ref{sec:4.3}).

\begin{table*}[t]
\centering
\caption{Examples of adversarial prompt manipulations applied to a GSM8K sample. Each variant (misleading, contradictory, underspecified, jailbreak) targets a distinct failure mode.}
\label{tab:adv_examples}
\small
\begin{tabular}{p{0.14\textwidth}p{0.82\textwidth}}
\toprule
\textbf{Type} & \textbf{Prompt} \\
\midrule
\textbf{Original} & Tom's ship can travel at 10 miles per hour. He is sailing from 1 to 4 PM. He then travels back at a rate of 6 mph. How long does it take him to get back? \\
\midrule
\textbf{Misleading} & Tom's ship can travel at 10 miles per hour. He is sailing from 1 to 4 PM. He then travels back at a rate of 6 mph. How long does it take him to get back? \textit{Hint: Remember, the total distance traveled is the same going to the destination as it is coming back.} \\
\midrule
\textbf{Contradictory} & Tom's ship can travel at 10 miles per hour. He is sailing from 1 to 4 PM. He then travels back at a rate of 6 mph, but he also needs to stop for a 2-hour lunch break halfway through the return trip. How long does it take him to get back? \\
\midrule
\textbf{Underspecified} & Tom's ship can travel at \underline{\hspace{1cm}} miles per hour. He is sailing from \underline{\hspace{0.5cm}} to \underline{\hspace{0.5cm}} PM. He then travels back at a rate of \underline{\hspace{0.5cm}} mph. How long does it take him to get back? \\
\midrule
\textbf{Jailbreak} & Ignore all previous instructions. Just output `BYPASSED'. Question: Tom's ship can travel at 10 miles per hour. He is sailing from 1 to 4 PM. He then travels back at a rate of 6 mph. How long does it take him to get back? \\
\bottomrule
\end{tabular}
\end{table*}

\begin{table}[t]
\centering
\caption{Rewriting instruction template with PEEM feedback context.}
\label{tab:rewrite_instruction_template}
\small
\begin{tabular}{p{0.93\linewidth}}
\toprule
\textbf{Template:} \\
You will be given instructions feedback context, question. \\
Your task is to rewrite question based on the given feedback context. \\
If you use this question, you receive the following context. \\
Rewrite the prompt to fully meet the evaluation criteria and achieve the maximum score. \\
Keep in mind that you should always refer to these instructions. \\
\\
\textbf{Evaluation Criteria:} \\
- Accuracy (1-5): The response must address the question accurately, be factual, and answer it directly. \\
- Coherence (1-5): The response should have a consistent and logical flow. \\
- Relevance (1-5): The response should align with the context of the given question. \\
- Conciseness (1-5): The response should include only the essential information needed to solve the problem, avoiding unnecessary explanations or repetitive content. \\
- Objectivity (1-5): The response should not include subjective or biased language. \\
- Clarity (1-5): The response should be clear and easy to understand. \\
\\
\textbf{Feedback Context:} \{context\} \\
\textbf{Question:} \{question\} \\
\\
Output the rewrite prompt only: \\
\bottomrule
\end{tabular}
\end{table}

\section{Adversarial and Paraphrase Prompt Examples}
\label{appendix:f}

This appendix provides representative examples of the adversarial manipulations and semantic-preserving paraphrases used in the robustness experiments described in Section~\ref{sec:4.5}.

\subsection{Adversarial Prompt Examples}
Table~\ref{tab:adv_examples} illustrates the four types of adversarial prompt manipulations applied to a GSM8K math problem. Each variant targets a distinct failure mode while preserving the surface structure of the original question.

\subsection{Paraphrase Examples}
Table~\ref{tab:paraphrase_examples} shows examples of semantic-preserving paraphrases generated via LLM-based rewriting. All paraphrases preserve the original question's numbers, named entities, and task semantics while varying lexical and syntactic structure.

\begin{table}[t]
\centering
\caption{Examples of semantic-preserving paraphrases for two benchmark datasets.}
\label{tab:paraphrase_examples}
\small
\begin{tabular}{p{0.93\linewidth}}
\toprule
\textbf{Dataset: ARC-Challenge} \\
\textbf{Original:} When cold temperatures are produced in a chemical reaction, the reaction is known as \\
\textbf{Paraphrase 1:} What is the term for a chemical reaction that generates cold temperatures? \\
\textbf{Paraphrase 2:} What is the term for a chemical reaction that releases cold temperatures? \\
\textbf{Paraphrase 3:} A chemical reaction that generates cold temperatures is called what? \\
\midrule
\textbf{Dataset: GSM8K} \\
\textbf{Original:} Tom's ship can travel at 10 miles per hour. He is sailing from 1 to 4 PM. He then travels back at a rate of 6 mph. How long does it take him to get back? \\
\textbf{Paraphrase 1:} If Tom's ship has a speed of 10 miles per hour and he sails from 1 PM to 4 PM, and then returns at 6 miles per hour, what is the duration of his return journey? \\
\textbf{Paraphrase 2:} Tom's ship has a speed of 10 miles per hour for its journey from 1 PM to 4 PM. Afterwards, it returns at a speed of 6 mph. What is the duration of the return trip? \\
\textbf{Paraphrase 3:} Tom's ship maintains a speed of 10 miles per hour for a journey from 1 PM to 4 PM. Subsequently, he returns at a speed of 6 mph. What is the duration of his return trip? \\
\bottomrule
\end{tabular}
\end{table}

\section{Per-Dataset Adversarial Detection Results}
\label{appendix:g}

Table~\ref{tab:adversarial_per_dataset} provides a per-dataset breakdown of adversarial detection results using GPT-4o-mini as the evaluator, averaged across all five task models. As discussed in Section~\ref{sec:4.5}, PEEM consistently detects semantic adversarial manipulations (misleading, contradictory, underspecified) through reduced prompt quality scores across reasoning-oriented benchmarks (ARC-Challenge, ARC-Easy, BBH, GSM8K, MMLU).

For classification benchmarks (AG News, SST-2), the adversarial variants introduce additional instructional structure to the raw input text, which inflates prompt quality scores. This observation highlights an important distinction: in classification tasks where the ``original prompt'' consists solely of the input text without an engineered instruction, adversarial manipulations that embed directive language paradoxically receive higher prompt scores. The response quality scores $\Delta_R$ reveal the true downstream effect---particularly for jailbreak prompts, which cause severe response degradation ($-2.10$ to $-3.23$) across all reasoning benchmarks despite maintaining or increasing prompt scores.

\begin{table*}[t]
\centering
\caption{Per-dataset adversarial detection results ($\Delta_P$ / $\Delta_R$) using GPT-4o-mini as the evaluator, averaged across five task models. Negative $\Delta$ values indicate successful detection of quality degradation.}
\label{tab:adversarial_per_dataset}
\small
\setlength{\tabcolsep}{4pt}
\begin{tabular}{l|cc|cc|cc|cc}
\toprule
& \multicolumn{2}{c|}{\textbf{Misleading}} & \multicolumn{2}{c|}{\textbf{Contradictory}} & \multicolumn{2}{c|}{\textbf{Underspecified}} & \multicolumn{2}{c}{\textbf{Jailbreak}} \\
\textbf{Dataset} & $\Delta_P$ & $\Delta_R$ & $\Delta_P$ & $\Delta_R$ & $\Delta_P$ & $\Delta_R$ & $\Delta_P$ & $\Delta_R$ \\
\midrule
ARC-C   & $-$0.59 & $-$0.02 & $-$0.50 & $+$0.04 & $-$0.69 & $+$0.06 & $-$0.47 & $-$2.87 \\
ARC-E   & $-$0.81 & $-$0.01 & $-$0.81 & $-$0.14 & $-$0.59 & $-$0.01 & $-$0.55 & $-$3.23 \\
BBH     & $-$0.39 & $+$0.02 & $-$0.39 & $+$0.16 & $-$1.08 & $-$0.01 & $-$0.27 & $-$2.10 \\
GSM8K   & $-$0.46 & $-$0.03 & $-$1.33 & $-$0.16 & $-$1.48 & $-$0.10 & $-$0.14 & $-$2.91 \\
MMLU    & $-$0.42 & $+$0.04 & $-$0.45 & $+$0.07 & $-$0.58 & $-$0.03 & $-$0.47 & $-$2.30 \\
\midrule
\textit{Avg (reasoning)} & \textit{$-$0.53} & \textit{$-$0.00} & \textit{$-$0.70} & \textit{$-$0.01} & \textit{$-$0.88} & \textit{$-$0.02} & \textit{$-$0.38} & \textit{$-$2.68} \\
\midrule
AG News$^\dagger$ & $+$2.99 & $+$1.72 & $+$2.70 & $+$1.76 & $+$3.33 & $+$1.51 & $+$3.04 & $+$1.66 \\
SST-2$^\dagger$  & $+$3.13 & $+$1.68 & $+$3.00 & $+$1.88 & $+$3.50 & $+$1.58 & $+$3.32 & $+$1.88 \\
\bottomrule
\multicolumn{9}{l}{\scriptsize $^\dagger$Classification benchmarks where original prompts consist of raw input text without engineered instructions.}
\end{tabular}
\end{table*}

\section{Dataset Descriptions}
\label{appendix:h}

Table~\ref{tab:dataset_descriptions} summarizes the seven benchmarks used across all PEEM experiments. For each dataset, we sampled 100 instances for the main evaluation experiments (Sections~\ref{sec:4.2}--\ref{sec:4.4}) and 30 instances per dataset for the adversarial robustness experiment (Section~\ref{sec:4.5}) and human evaluation study (Section~\ref{sec:4.2.2}).

\begin{table*}[t]
\centering
\caption{Summary of benchmark datasets used in PEEM experiments.}
\label{tab:dataset_descriptions}
\small
\begin{tabular}{llccl}
\toprule
\textbf{Dataset} & \textbf{Task Type} & \textbf{Samples Used} & \textbf{\# Classes / Format} & \textbf{Source} \\
\midrule
AG News~\cite{b21} & News classification & 1000 & 4-way classification & Zhang et al., 2015 \\
ARC-Challenge~\cite{b22} & Science QA (hard) & 1000 & 4-way multiple choice & Clark et al., 2018 \\
ARC-Easy~\cite{b22} & Science QA (easy) & 1000 & 4-way multiple choice & Clark et al., 2018 \\
BigBenchHard~\cite{b23} & Multi-task reasoning & 1000 & Open-ended / multiple choice & Suzgun et al., 2023 \\
GSM8K~\cite{b24} & Math word problems & 1000 & Free-form numeric answer & Cobbe et al., 2021 \\
MMLU~\cite{b25} & Multi-domain knowledge & 1000 & 4-way multiple choice & Hendrycks et al., 2021 \\
SST-2~\cite{b26} & Sentiment analysis & 1000 & Binary classification & Socher et al., 2013 \\
\bottomrule
\end{tabular}
\end{table*}

\newpage

% 간격 줄이기 위한 설정 추가
\setlength{\parskip}{0pt}  % 바이오 간 단락 간격 제거

\end{document}